\documentclass{jmlr}

\usepackage{longtable}

\usepackage{booktabs}
\usepackage[load-configurations=version-1]{siunitx} 

\theorembodyfont{\upshape}
\theoremheaderfont{\scshape}
\theorempostheader{:}
\theoremsep{\newline}

\title[Geometric Representational Alignment]{Geometry matters: insights from Ollivier Ricci Curvature and Ricci Flow into representational alignment}

\author{\Name{Nahid Torbati} \Email{torbati@cbs.mpg.de}\\
\addr Max Planck Institute for Human Cognitive and Brain Sciences, Leipzig, Germany
\AND
\Name{Michael Gaebler} \Email{gaebler@cbs.mpg.de}\\
\addr Max Planck Institute for Human Cognitive and Brain Sciences, Leipzig, Germany\\
\addr Berlin School of Mind and Brain, Humboldt-Universität zu Berlin, Berlin, Germany
\AND
\Name{Simon M. Hofmann} \Email{simon.hofmann@cbs.mpg.de}\\
\addr Max Planck Institute for Human Cognitive and Brain Sciences, Leipzig, Germany
\AND
\Name{Nico Scherf} \Email{nscherf@cbs.mpg.de}\\
\addr Max Planck Institute for Human Cognitive and Brain Sciences, Leipzig, Germany\\
\addr Center for Scalable Data Analytics and Artificial Intelligence (ScaDS.AI), Dresden/Leipzig, Germany\\
}

\usepackage{etoolbox} 

\makeatletter
\providecommand\origmarkboth{}%
\let\origmarkboth\markboth
\newcommand{\suppressjmlrmarks}{\let\markboth\@gobbletwo} 
\newcommand{\restorejmlrmarks}{\let\markboth\origmarkboth}

\AtBeginDocument{%
  \let\oldmaketitle\maketitle
  \renewcommand{\maketitle}{%
    \suppressjmlrmarks
    \oldmaketitle
    \restorejmlrmarks
    \markboth{}{}
    \thispagestyle{plain}
  }%
}
\makeatother

\begin{document}
\maketitle
\begin{abstract}

Representational similarity analysis (RSA) is widely used to analyze the alignment between humans and neural networks; however, conclusions based on this approach can be misleading without considering the underlying representational geometry.
Our work introduces a framework using Ollivier Ricci Curvature and Ricci Flow to analyze the fine-grained local structure of representations. This approach is agnostic to the source of the representational space, enabling a direct geometric comparison between human behavioral judgments and a model's vector embeddings. We apply it to compare human similarity judgments for 2D and 3D face stimuli with a baseline 2D-native network (VGG-Face) and a variant of it aligned to human behavior.
Our results suggest that geometry-aware analysis provides a more sensitive characterization of discrepancies and geometric dissimilarities in the underlying representations that remain only partially captured by RSA.
Notably, we reveal geometric inconsistencies in the alignment when moving from 2D to 3D viewing conditions.
This highlights how incorporating geometric information can expose alignment differences missed by traditional metrics, offering deeper insight into representational organization.

\end{abstract}
\begin{keywords}
Ollivier Ricci Curvature, Ricci Flow, Representational Alignment, Convolutional Neural Networks
\end{keywords}

\section{Introduction}
\label{sec:intro}
Understanding how biological and artificial neural networks (ANNs) solve common tasks such as face recognition or object classification in natural scenes remains a central challenge at the intersection of cognitive science and machine learning. A key question is whether these systems, despite their architectural and physical differences among others, develop similar internal representations when faced with the same computational demands. This has given rise to the study of \textit{representational alignment}, which seeks to uncover not only how much representations across systems correspond, but also \textit{how} and \textit{why} such (dis)similarities emerge \citep{sucholutsky2023getting}, ultimately exploring both universal and idiosyncratic aspects of representations across systems. 
Common methods to behaviorally probe the representational space in humans and computational models use perceptual downstream tasks such as similarity judgments (e.g., in \textit{odd-one-out} designs) between objects \citep{mahner2025dimensions} or faces \citep{hofmann2024human}.
\textit{Representational Similarity Analysis} (RSA) is a multivariate method for comparing representational structures across modalities, especially between biological and artificial systems \citep{kriegeskorte2008representational}. By correlating Representational Dissimilarity Matrices (RDMs) or similarity kernels, RSA captures second-order isomorphisms, comparing the relational structure of stimulus representations, rather than directly mapping activity patterns. Despite its popularity, RSA is limited by its reliance on Euclidean geometry, which can distort the intrinsic structure of representational spaces. Empirical work further shows that human similarity judgments often deviate from Euclidean assumptions \citep{rodriguez2017differential}, restricting RSA’s ability to capture alignment grounded in the true geometry of the data.\\
While RSA is widely used, multiple studies highlight critical limitations. Geometrical similarity does not guarantee comparable representational content, as different coding schemes can yield similar RDM structures \citep{laakso2000content}. Dataset and feature dependencies can inflate similarity, such as between texture-based DNNs and shape-based human vision \citep{dujmovic2022some}. Diagonal entries in RDMs can artificially boost correlations \citep{ritchie2017avoiding}, and dissimilarity estimates suffer from bias, covariance, and noise sensitivity, prompting methods such as multivariate noise normalization, cross-validation, and unbiased estimators \citep{walther2016reliability,diedrichsen2020comparing}. Moreover, human similarity judgments often deviate from Euclidean assumptions, aligning better with Riemannian geometry \citep{rodriguez2017differential} and violating symmetry and triangle inequality constraints \citep{tversky1977features,tversky1982similarity}. Finally, estimation biases tied to task design and within-run correlations further undermine standard RSA, motivating Bayesian approaches that model raw data directly \citep{cai2019representational}.\\
To address the limitations of RSA and capture the intrinsic representational geometry, we model each space as a graph, a discrete analog of a manifold, and apply Ollivier–Ricci curvature (ORC) \citep{ollivier2007ricci} to quantify local connectivity via optimal transport between neighborhood probability measures.
 These analyses enabled us to quantify both local and global (dis)similarities between representations. In the human dataset, we further compared 2D and 3D viewing conditions, examining their representational congruence and how it is embedded in the underlying geometry. 

\paragraph{Main contributions}
\begin{enumerate}
   \item \textbf{A novel geometry-aware framework}: We introduce an embedding-agnostic framework for representational alignment using Ollivier Ricci Curvature and Ricci Flow. This approach moves beyond the Euclidean assumptions of traditional Representational Similarity Analysis (RSA) to capture and quantify the intrinsic local geometry and community structure of representational spaces.

    \item \textbf{Revealing the limits of behavioral alignment}: We demonstrate that successfully aligning a network to human choices does not guarantee an alignment of the underlying representational geometry, especially under a modality mismatch: We train a 2D-native network (VGG-Face) on human similarity judgments derived from 3D stimuli and find that while task performance is high, the network fails to reconstruct the geometric complexity of human 3D perception. Our method reveals this failure, whereas traditional metrics suggest a moderate alignment.

    \item \textbf{Identifying alignment deficits to form relevant semantic structures}: Using our framework, we show the network's inability to form key semantic structures, here, male and female clusters,
    suggesting limitations of the 2D-native model to align its representational space with the human behavioral data from the richer dynamic 3D experience.

\end{enumerate}

\section{Related work}
\paragraph{Representational alignment}
\citet{sucholutsky2023getting} explore the challenging concept of representational alignment, which assesses the degree to which internal representations between different information processing systems - biological and artificial - correspond.
A central approach to bridge neuroscientific domains is Representational Similarity Analysis (RSA), which allows for the comparison of representational structures across measurement modalities \cite{kriegeskorte2008representational}. RSA contrasts how brain regions encode information using representational differences matrices (RDM) to capture differences in neural responses between stimuli.
RSA has been widely used to compare artificial and biological systems, evaluating how well computational models replicate the representational geometry of processes across brain regions. Studies have employed RSA to contrast supervised and unsupervised learning models and to assess the impact of model scale, architecture, training data, and objective functions on alignment with neural representations \citet{khaligh2014deep,yamins2016using, muttenthaler2022human,sucholutsky2024alignment}.
Beyond RSA, a variety of complementary techniques have been developed to investigate neural network representations. These range from global similarity metrics based on kernel alignment \citep{kornblith2019similarity}, to finer-grained methods that assess the preservation of local neighborhood structure \citep{kolling2023pointwise} \cite{huh2024platonic}. Other approaches leverage geometric principles, such as anchor-based encodings \citep{moschella2022relative} \citep{cannistraci2023bricks} or spectral maps \citep{fumero2024latent}, to enable invariant comparisons between representational spaces. This body of work reflects a trend towards capturing more nuanced structural properties beyond what is revealed by a single global similarity score.
\cite{huh2024platonic} use a mutual nearest-neighbor (mNN) metric to measure how often two models share the same top \textit{k} neighbors per sample, capturing local similarity better than global metrics like Centered Kernel Alignment (CKA; \citet{kornblith2019similarity}). Applying mNN across models and modalities, it shows converging representations driven by scale, task diversity, and simplicity bias, supporting the idea of a shared "Platonic" representation in AI.
\paragraph{Ollivier Ricci Curvature and Ricci Flow: Tools for structural analysis}
The classical notion of Ricci curvature from Riemannian geometry was extended to discrete settings by Ollivier through an optimal transport framework \citep{ollivier2007ricci}, giving rise to Ollivier–Ricci curvature (ORC) on graphs and networks. Hamilton’s Ricci flow \citep{hamilton1982three}, originally formulated as the geometric evolution of curvature, analogous to heat diffusion, was later discretized and adapted to graphs and networks by \citet{ni2019community}.
Together, these concepts provide a geometric lens for analyzing complex network structures. The discretized Ricci flow has been employed as a principled approach for community detection and network alignment \citep{ni2019community,ni2018network}. ORC itself has been widely adopted across diverse domains: for instance, \citet{topping2021understanding} introduced Balanced Forman curvature, inspired by ORC, to characterize the "over-squashing" phenomenon. 
Extensions of ORC further by incorporating dynamical similarities, revealing how curvature distributions evolve and identify bottleneck edges \citep{topping2021understanding}. In weighted graphs, it has been used to quantify the local "bending" or geometric distortion at individual edges \citep{sandhu2015graph}. In finance, it provides a novel lens to evaluate systemic risk and market fragility \citep{sandhu2016ricci}.
In neuroscience, ORC has proven effective in tracking age-related changes in brain networks as well as connectivity alterations associated with autism spectrum disorder, uncovering both region-specific and global patterns of structural and functional differences \citep{farooq2019network,elumalai2022graph,yadav2023discrete}.

\section{Methods}
\label{sec:methods}

\paragraph{Ricci curvature, Ollivier Ricci curvature}
In Riemannian geometry, curvature measures how a manifold deviates from being locally Euclidean, with Ricci curvature specifically quantifying this deviation in tangent directions \citet{riemann2016hypotheses,samal2018comparative}. It influences the average spread of geodesics in those directions and the rate at which the volume of distance balls and spheres expands.


Ricci curvature affects how the volume of a ball grows with its radius and the overlap between two balls, which depends on their radii and the distance between centers. Greater overlap implies lower transport cost, linking Ricci curvature to optimal transport. Building on this idea, Ollivier introduced a generalized Ricci curvature for metric measure spaces based on optimal transport \citet{ollivier2007ricci,ollivier2009ricci}. For a metric space $(X, d)$ equipped with a probability measure $m_x$ for each $x\in X$, the Ollivier Ricci curvature (ORC) $\kappa_{xy}$ along a path $xy$ is defined as follows:

\begin{equation}\label{eq:ollivier}
    \kappa_{xy}=1-\frac{W(m_x,m_y)}{d(x,y)}
\end{equation}

where $W(m_x,m_y)$ is the Wasserstein distance. ORC provides an edge-based measure of local connectivity in graphs. Positive curvature indicates cohesive neighborhoods with low transport cost, while negative curvature highlights structural bottlenecks or sparse connections. Intuitively, ORC captures how easily “mass” flows across edges, distinguishing well-connected regions from areas of separation.

\paragraph{Analysis of curvature distributions.} To quantitatively characterize the shape of the Ollivier Ricci Curvature distributions from each graph, we modeled them using Gaussian Mixture Models (GMMs). This allows us to identify whether a distribution is simple and unimodal or complex and multimodal. The models were implemented using the scikit-learn (version 1.3.2) library \cite{scikit-learn} in Python. In addition, we applied the Wasserstein distance to measure the difference between the curvature distributions, more precisely, Wasserstein 1-distance-1 ($WS_1$). For the computation, the library SciPy (version 1.10.1) in Python has been used \cite{2020SciPy-NMeth}.

\paragraph{Ricci flow and Ricci flow-metric}
\label{par:ricciflow}
The Ricci flow method, based on the geometric concept of curvature introduced by F. Gauss and B. Riemann, describes how the space bends at each point \citep{perelman2002entropy,gauss1828disquisitiones,riemann2016hypotheses}. Areas with high positive curvature are denser, while regions with negative curvature are less so. Hamilton developed the Ricci flow, a curvature-driven diffusion process, which deforms space similarly to heat diffusion \citep{hamilton1982three}. 
\citet{ni2019community} adapted Ricci flow from Riemannian geometry to discrete networks, using it to detect community structures within graphs. The discrete Ricci flow algorithm on a network is an evolving process; in each iteration, all edge weights are updated simultaneously by the following flow process:

\begin{equation}
\label{for:flow}
    w_{xy}^{i+1} = d^i(x,y)-\kappa^i(x,y) \cdot d^i(x,y)
\end{equation}

where $w^i_{xy}$ is the weight of the edge $xy$ in the $i$-th iteration, and $\kappa_{xy}^i$ is the curvature value at the edge $xy$ in the $i$-th iteration, and $d^i_{(x, y)}$ is the shortest path distance in the graph induced by the weights $w_{xy}^i$. \\
\textbf{Ricci flow-metric}, the Ricci flow-induced distance is defined based on the definition of Ricci flow
\citep{ni2018network}. It is calculated between nodes using the weighted shortest path, where the edge weights are derived from the final output of the flow process. Details of computations and settings for the further analysis are provided in Appendix \ref{apd:second}.

\paragraph{Heat diffusion distance}
To compare graphs based on edge curvature, we use the graph diffusion distance \citep{hammond2013graph}, which quantifies the average similarity of heat diffusion patterns and enables comparison between weighted graphs. 
Diffusion is simulated by initiating a delta impulse at a vertex and allowing it to evolve over time $t$; differences in adjacency matrices produce distinct diffusion patterns. 
The distance between two graphs $G_1$ and $G_2$ and for time $t$ can be calculated with:

\begin{equation}
\label{eq:GDD}
    d(G_1,G_2;t) = \|exp (-tL_1) - exp(-tL_2)\|_F^2
\end{equation}

where $\|.\|_F$ is the Frobenius norm and $L_1$ and $L_2$ are Laplacian matrices of $G_1$ and $G_2$, respectively.


\paragraph{Human dataset}
The human behavioral data (\textbf{Human Judgment}) were obtained from a large-scale online experiment \citep{hofmann2024human}. 2,710 participants judged similarities among 100 generated faces (3D head models; $n_{\text{female}}=n_{\text{male}}=50$; 161,700 trials) in a triplet odd-one-out task under static 2D and dynamic 3D viewing conditions. In the 2D condition, 3D head models (faces) were shown statically from a frontal perspective; in the 3D viewing condition (or 2.5D condition; \citep{marrVisionComputationalInvestigation1982}) the head-models were continuously rotated such that participants could integrate various perspectives in their similarity judgments.

\paragraph{General approach}
Human judgments were modeled with a custom-aligned \textbf{VGG-Face}, \citep{parkhi2015deep}, which we call \textbf{Aligned-VGG}, trained to predict human odd-one-out choices in triplets of face images. 
Similarity judgments are expressed as similarity matrices. From these, we construct a unified graph-based representation using \textit{adaptive-}KNN graphs for both synthetically generated datasets (\figureref{fig:syn_data}, Appendix~\ref{apd:second}) and model representational spaces, with construction details given in Appendix~\ref{apd:second}. We then compute Ollivier–Ricci curvature (ORC) for each representation (Human Judgment, Aligned-VGG, and VGG-Face) and apply the discrete Ricci flow algorithm. The resulting flow-metric is used to analyze the representations and their underlying geometric properties.

\section{Results}
\paragraph{Human-like geometry emerges in 2D viewing conditions but generalizes weakly to 3D.}
To investigate how behavioral alignment shapes representational geometry, we compared human judgments with the aligned and baseline networks across both 2D and 3D viewing conditions (\figureref{fig:alignment_heatmap}). While a human-like geometry is successfully induced in the 2D condition, our analysis shows that this representational alignment is much weaker under the 3D input modality, despite behavioral alignment being comparably strong across conditions (performance scores of Aligned-VGG 2D: 0.72, and 3D: 0.69; noise-ceiling corrected; for details, see \citet{hofmann2024human}).
In the \textbf{2D condition} (\figureref{fig:alignment_heatmap}a), the RDMs for both Human Judgment (left) and the Aligned-VGG network (middle) show a clear block-diagonal structure when using our flow-metric, indicating a shared geometric organization based on gender. This visual similarity is confirmed by the high quantitative correlation between them (r=0.63). This result establishes that, under matched conditions, behavioral alignment can indeed induce a human-like representational geometry.


\begin{figure}[h]
\floatconts
  {fig:alignment_heatmap}
  {\caption{Representational dissimilarity matrices (RDMs) and corresponding alignment scores for (a) 2D and (b) 3D conditions. To compare traditional methods with our geometric approach, we compute RDMs using three distance metrics: \textit{Euclidean} (a standard baseline), \textit{cosine} (a strong empirical baseline \cite{diedrichsen2020comparing}), and our novel \textit{flow-metric}, which captures intrinsic geometry. Alignment scores (top of each panel) are the Pearson correlation between respective RDMs. The first 50 images (e.g., rows) correspond to female faces and the next 50 images to male faces, revealing a gender-based block structure.}}
  {\includegraphics[width=0.9\linewidth]{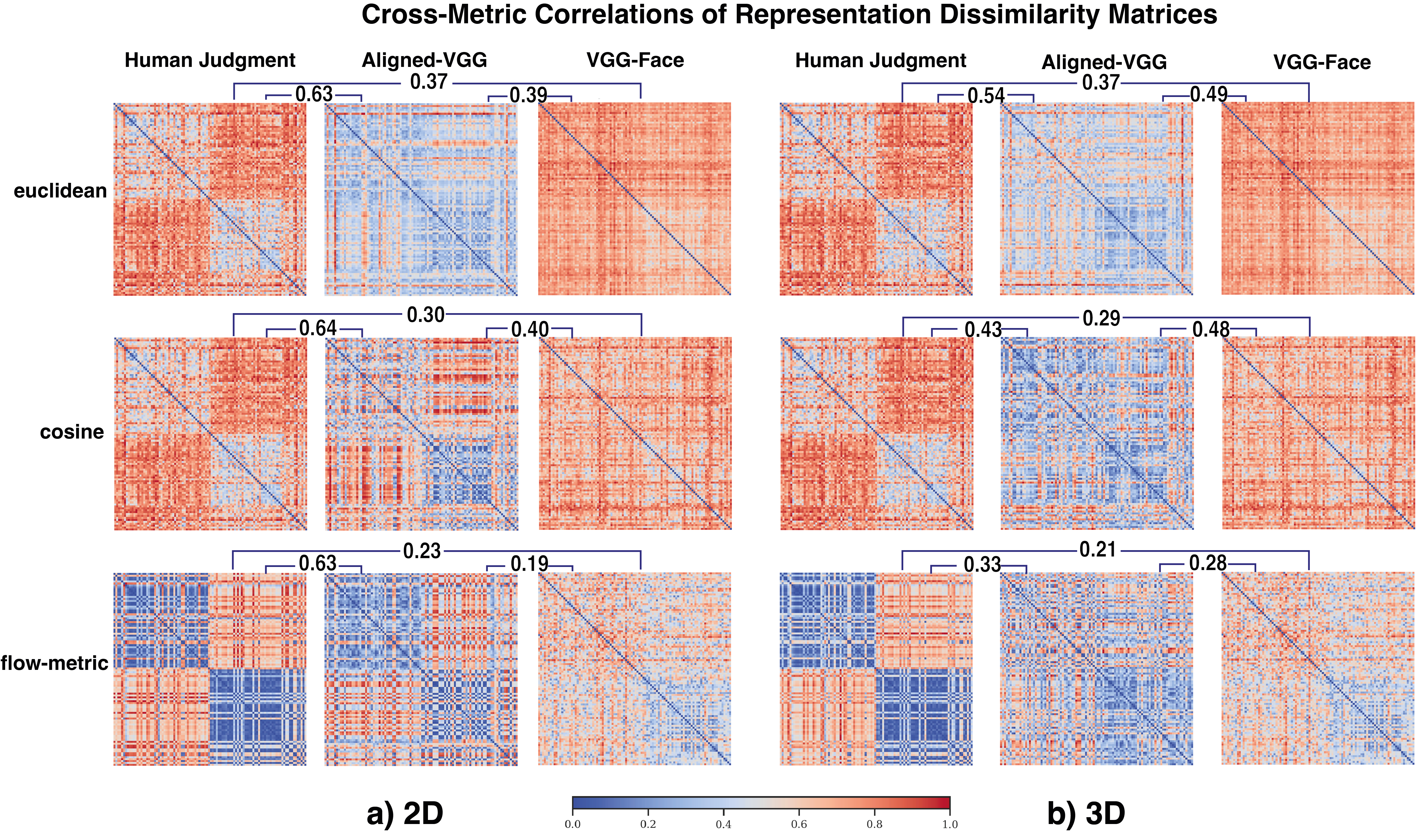}}
\end{figure}

However, this alignment does not readily generalize when the network is aligned to behavior from the \textbf{3D viewing condition} (\figureref{fig:alignment_heatmap}b). The clear geometric structure in the Aligned-VGG model disappears; its flow-metric RDM (middle column, third row) now lacks the distinct block structure seen in both the 2D case and in the 3D Human Judgment RDM.

Critically, this reduced matching of geometric structure is largely missed by traditional metrics. While the standard Euclidean and Cosine correlations between Human Judgment and Aligned-VGG remain moderate (r=0.54 and r=0.43, respectively), the flow-metric correlation reduces to 0.33. This demonstrates that while the network can still match human choices at a surface level, it doesn't fully reconstruct the underlying geometry of human perception, a misalignment that our geometry-aware method reveals. This change in alignment is specifically tied to the semantic structure of the faces, as the network loses the clear separation of the female face cluster (first 50 images) that is prominent in the human data.

\paragraph{Curvature distributions reveal geometric misalignment.}

To understand the geometric source of the reduced alignment described above, we analyzed the Ollivier Ricci Curvature (ORC) distributions for each representation using three complementary tools: Gaussian Mixture Models (to identify structural complexity in the distribution), Wasserstein distance (to measure the (dis)similarity of distributions), and heat diffusion distance (to measure global graph (dis)similarity including geometrical information; see Methods for details). This analysis confirms the 2D/3D divergence: the geometry of the Aligned-VGG network mirrors human judgments in the 2D condition but regresses toward the simpler geometry of the baseline VGG-Face in the 3D condition (see Methods, \tableref{tab:statistical-anal} and \tableref{tab:gaussian-results} in Appendix \ref{apd:second}).


In the \textbf{2D condition} (\tableref{tab:statistical-anal}) the curvature profiles of Human Judgment and Aligned-VGG are highly similar, reflected by a low Wasserstein distance between their distributions (0.04) and a strong correspondence in their heat diffusion patterns (1.6). Both representations exhibit a complex, multi-modal curvature distribution, suggesting a nuanced geometric structure (see \figureref{fig:Gaussians-image-2D} and \tableref{tab:gaussian-results} in  Appendix \ref{apd:second}).


In contrast, in the \textbf{3D condition}, the geometry of the Aligned-VGG model does not show human-like complexity and remains statistically closer to the baseline VGG-Face. Both networks fit a simple, single-Gaussian curvature distribution (\tableref{tab:gaussian-results}), and the Wasserstein and heat diffusion distances between them are smaller than their distances to Human Judgment(\tableref{tab:statistical-anal}). In contrast, the Human Judgment representation is geometrically more complex, with two distinct components in its curvature distribution (\tableref{tab:gaussian-results}). 

This indicates that the behavioral signal from the dynamic 3D viewing condition, when fed to the 2D-native model, was insufficient to impose the complex geometric structure of human perception. This critical geometric separation is robust across a range of graph construction parameters. Increasing connectivity parameters modestly improves the alignment score only between Aligned-VGG and VGG-Face, while Human Judgment remains consistently separate (\figureref{fig:heat-distances-2D} and \figureref{fig:ws-distances-2D} in Appendix \ref{apd:second}).

\begin{table}[h]
\floatconts
  {tab:statistical-anal}
  {\caption{Representational comparison across models in 2D and 3D conditions for HJ (Human Judgment), A-VGG (Aligned-VGG), and VGG-Face. Values are the mean and standard deviation of the Wasserstein (WS) and heat diffusion distances, computed over 100 iterations on random subsets of 70 data points to ensure statistical robustness.}}
  {\begin{tabular}{lccc}
  \toprule
  \bfseries \textbf{Model} & \textbf{HJ vs. A-VGG} & \textbf{HJ vs. VGG-Face} & \textbf{A-VGG vs. VGG-Face}\\
    \midrule
    WS-distance 2D    & 0.04 $\pm$ 0.01 & 0.12 $\pm$ 0.01 & 0.12 $\pm$ 0.01\\
    WS-distance 3D      & 0.07 $\pm$ 0.01 & 0.13 $\pm$ 0.01 & 0.07 $\pm$ 0.01\\
    Heat-distance 2D   & 1.6 $\pm$ 0.04 & 2.05 $\pm$ 0.03 & 1.98 $\pm$ 0.03\\
    Heat-distance 3D   & 1.69 $\pm$ 0.04 & 2.08 $\pm$ 0.04\ & 1.79 $\pm$ 0.03\\
  \bottomrule
\end{tabular}}
\end{table}

\paragraph{Community structure reveals divergent alignment}
\label{subsec:ricciflow}

Finally, to understand how these local geometric differences give rise to meso-scale organization, we used discrete Ricci flow \citep{ni2019community}. Ricci flow is a curvature-driven process that contracts well-connected regions and expands sparse ones, thereby it provides a principled way to detect emerging community structure in the graphs (\figureref{fig:ricci-flow}). The results confirm the pattern of 2D alignment and the reduced alignment in 3D. In the 2D condition (\figureref{fig:ricci-flow}a), the Human Judgment and Aligned-VGG graphs exhibit similar community patterns, both showing a clear tendency to cluster faces by gender. In contrast, the baseline VGG-Face model lacks a coherent structure.

This structure is less pronounced in the model variant trained on data of the 3D condition (\figureref{fig:ricci-flow}b). Although the community structure in human judgments also shifts between the 2D and 3D conditions (e.g., by the number of clusters), the Aligned-VGG model fails to capture this organization. Instead, its graph more closely resembles the baseline VGG-Face, a finding supported by quantitative community metrics (see Appendix \ref{apd:second} and \tableref{tab:results}).

By tracking representative faces across the graphs, we can get visual insights into the specific semantic structure that does not generalize (\figureref{fig:ricci-flow}), indicating that the emergence of distinct clusters in the 3D Aligned-VGG is more limited compared to the 2D case. 
This is exemplified in clusters of female face images, which are successfully formed during 2D alignment, however, appear incoherent in the 3D condition, highlighting the network’s failure to learn the geometry of human perception under a modality mismatch (see Appendix \ref{apd:second} for a fine-grained Representational Profile Analysis, which confirms that this geometric divergence is driven by the incomplete emergence of the female face cluster; see \figureref{fig:row-wise-3D} and \figureref{fig:row-wise-2D}).

\begin{figure}[h]
\floatconts
  {fig:ricci-flow}
  {\caption{
  Communities (different symbols) detected by Ricci flow in 2D and 3D viewing conditions. To visualize cluster coherence, representative faces were selected from the communities in the 3D Human Judgment graph and are shown across all panels. Edge colors indicate curvature (red=negative, blue=positive), where red edges highlight intra-community relations; their absence in the 3D Aligned-VGG graph reveals a lack of coherent cluster structure. The GMM component values of the curvature distributions are also shown in the distribution of curvature values.
  }}
  {\includegraphics[width=0.9\linewidth]{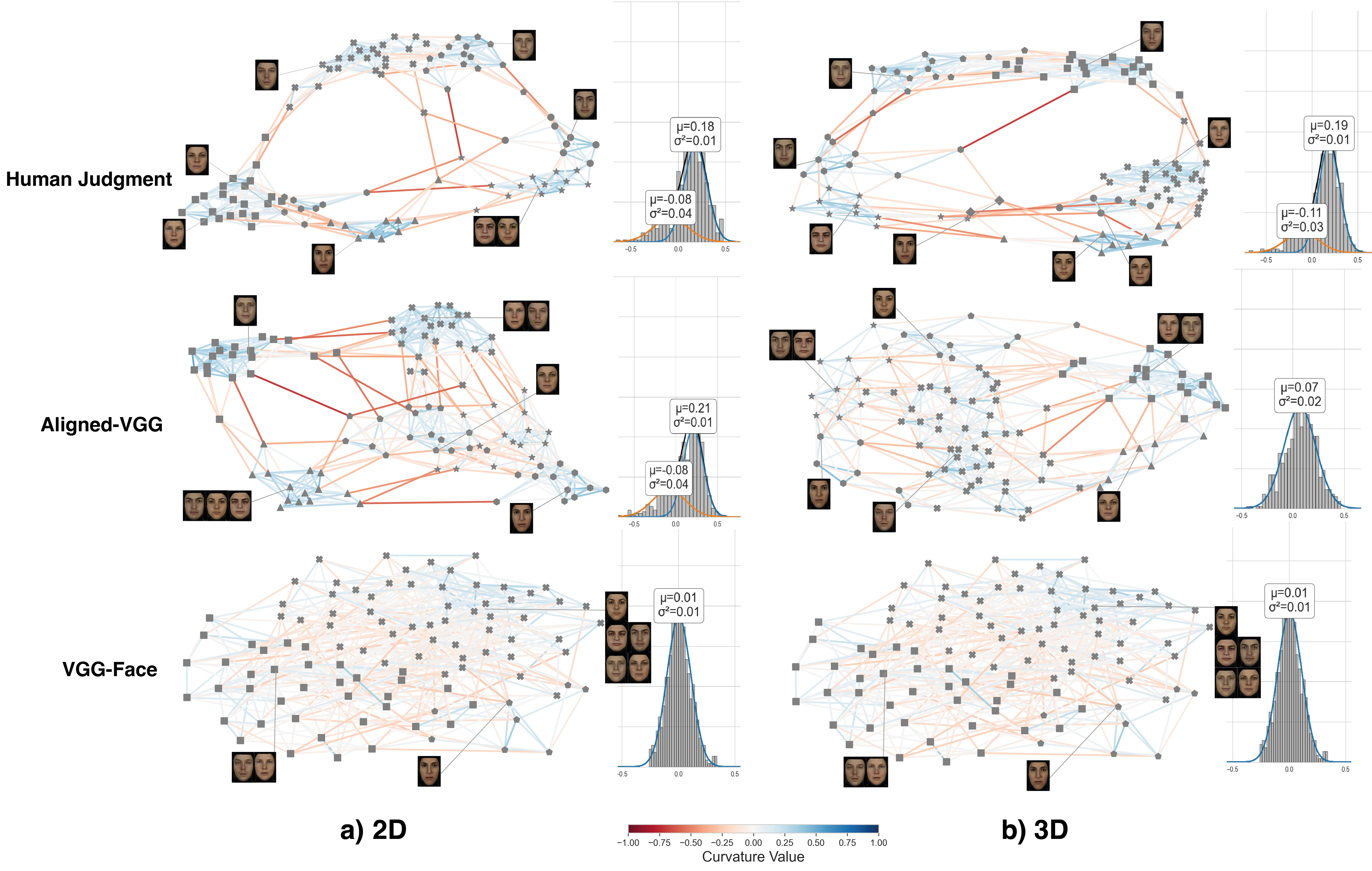}}
\end{figure}


\section{Discussion and Conclusions}
\label{sec:discussion}

In summary, our study yields three key findings: First, by employing a novel geometry-aware framework that leverages curvature to incorporate local geometric information when measuring alignment between representations, we show that traditional metrics can miss important discrepancies in the underlying representational geometry. Thereby, our approach reveals inherent limitations of behavioral alignment , that is, high task performance does not guarantee a human-like internal organization. Second, it demonstrates that the geometric consequences of changing alignment conditions  (here, from 2D to 3D)–while task-performance remains relatively constant–are not fully captured by the standard distant metrics. Third, our framework uncovers a model's incapacity to form a coherent semantic structure (here, for female face images in the 3D viewing condition).
More specifically, our results indicate that behavioral feedback in the dynamic 3D condition imposes weaker constraints on the model's representational geometry compared to the static 2D case. That is, even when guided by human judgments, the 2D-native network (VGG-Face) fails to converge to a geometrically similar representation. This suggests that successful alignment requires the model’s representational capacities (e.g., processing 3D information) to match the complexity of the human experience it is intended to capture.


Our findings also point to important directions for future work. First, our analysis was limited to a single model family (VGG-Face);  an important next step is to test whether these geometric findings generalize to other architectures, such as Transformers \citep{vaswani2017attention}. Second, our study relied on controlled, computer-generated faces; examining whether the results hold for more naturalistic images with greater variability will be essential for establishing their broader relevance. Third, while our graph-based approach is robust, it depends on discrete approximations of curvature; exploring alternative formulations of these geometric concepts could provide deeper insights. Finally, moving beyond behavioral alignment, future work could develop geometrically guided alignment methods that incorporate elements of our proposed framework.

To conclude, challenging the limits of classic similarity metrics, our work highlights the potential of discrete geometric tools to provide a more rigorous and insightful lens for comparing perception and representation across biological and artificial systems.

\acks{N.S., N.T, and S.H. are supported by BMBF (Federal Ministry of Education and Research) through ACONITE (01IS22065) and the Center for Scalable Data Analytics and Artificial Intelligence (ScaDS.AI.) Dresden/Leipzig. N.T. is also supported by the Max Planck IMPRS CoNI Doctoral Program.\\
M.G. and S.H. were funded by a cooperation between the Max Planck Society and the Fraunhofer-Gesellschaft (project NEUROHUM).}

\bibliography{pmlr-sample}

\appendix

\section{Human experiment data}\label{apd:data}
\paragraph{Face stimuli and human similarity judgments}
Stimulus images were computed using the 3D reconstruction model DECA \citep{feng2021learning} applied to 2D portraits of the Chicago Face Dataset \citep{maChicagoFaceDatabase2015}. Human face similarity judgments (n = 194,261) were acquired in the form of a triplet-odd-one-out task from 2,710 participants (age range 18 - 65, mean age = 31.9 ± 11.2 years) in an online experiment. For more details on the stimulus set and experimental design, we refer to \citet{hofmann2024human}.

\begin{figure}[t]
     \centering
     \includegraphics[width=0.55\linewidth]{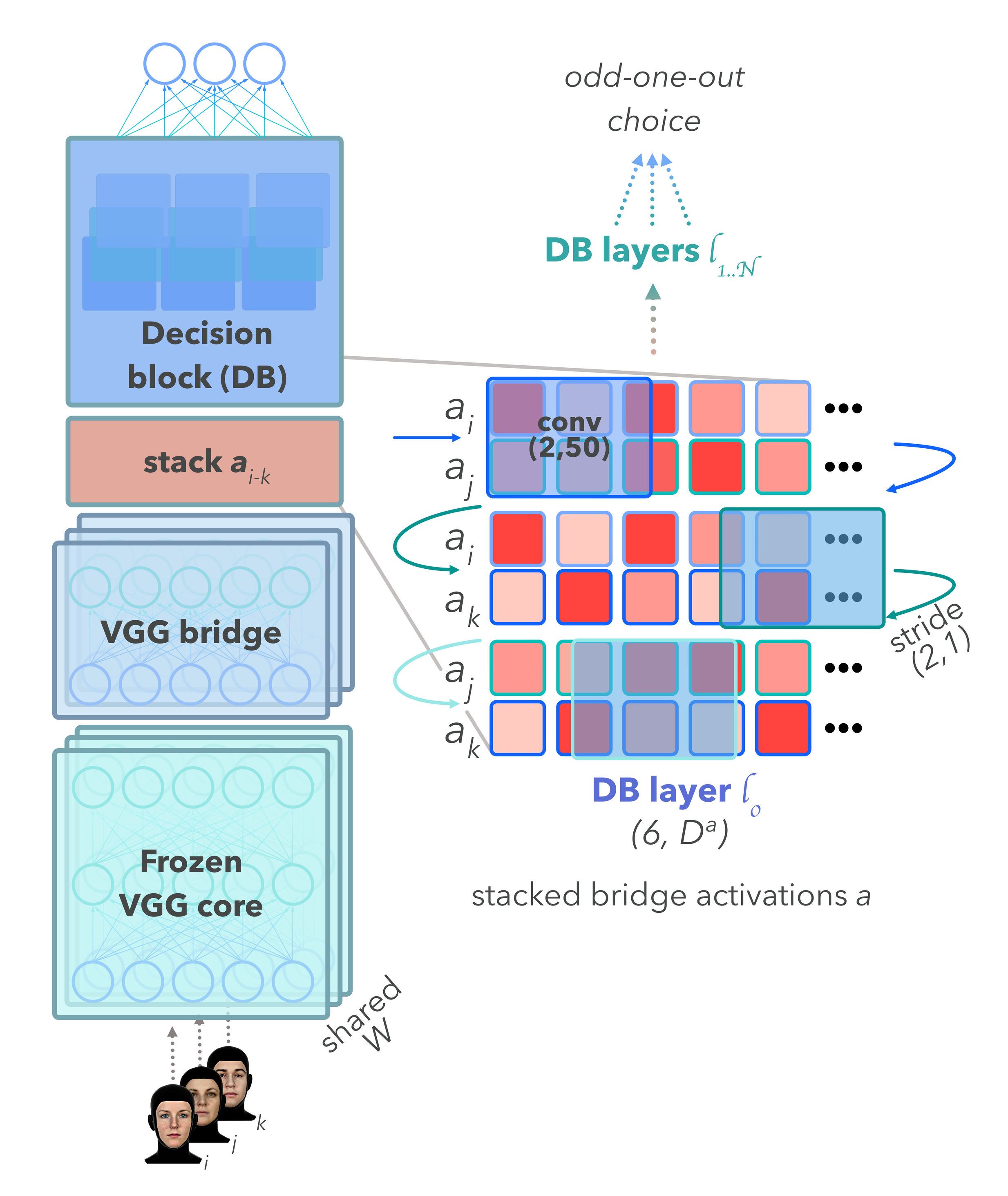}
     \caption{\textbf{Human-aligned VGG-Face}. The network is trained to predict human judgments in a face similarity task.}
     \label{fig:aligned-VGG-model}
\end{figure}

\paragraph{Human-aligned VGG-Face}
\label{apd:aligned-vgg-face}
The pre-trained VGG-Face architecture \citep{parkhi2015deep} was adopted to predict human face similarity judgments in the experiment (Figure \ref{fig:aligned-VGG-model}). 
First, all layers up to the fully connected layer FC7 were frozen, making their weights non-trainable (\textit{VGG core}). Second, subsequent layers were replaced with one FC layer (\textit{VGG bridge}), which converts a 4,096-dimensional input to a 300-dimensional vector. A \textit{decision block} was added, consisting of convolutional layers. This block receives stacked activation maps from the bridge for each input image in a triplet $(x_i, x_j, x_k)$, resulting in a 6x300 matrix $[a_i, a_j, a_i, a_k, a_j, a_k]$. The first convolutional layer in the decision block has 2 filters of size (2, 50) and stride (2, 1), producing an output of size (batch size, 2, 3, 251). After applying a ReLU activation, another convolutional layer with one filter of size (3, 100) and stride (1, 1) is applied, followed by another ReLU. This results in a (batch size, 2, 1, 152) output.
Then, the signal was down-sampled to (batch size, 1, 1, 3) using two more convolutional layers (one filter each, kernel sizes: (1, 100) and (1, 51)) with an intermediate ReLU. The resulting 3-length output vector indicates the model's choice, where the highest value identifies the odd-one-out.
The architecture was trained using cross-entropy loss with the Adam optimizer, a learning rate of 5e\textsuperscript{-4}, and a batch size of 16. The data (X: triplet images, Y: human choices) was split into training (70\%; $n_{train} = 135,982$), validation (15\%; $n_{val} = 29,139$), and test sets (15\%; $n_{test} = 29,140$). For more details, see \citet{hofmann2024human}.

\section{}\label{apd:second}
\subsection{Graph construction}
\label{par:Graph}
To construct the graph from vector embeddings, we employ an adaptive nearest-neighbor method. First, we calculate the density as the inverse of the distance using a k-nearest neighbors (KNN) density kernel, incorporating the parameters $k_{min}$ and $k_{max}$, which specify the minimum and maximum number of neighbors, respectively. Next, we normalize the local density by scaling it between the minimum and maximum density values, thereby defining the density at each data point. Based on this normalized density, we determine the number of neighbors for each data point and proceed to construct the graph. The same approach is applied to the similarity matrix derived from human similarity judgments. First, a distance matrix is constructed from the similarity matrix provided. Using this distance matrix, the method estimates an appropriate $k$-value for each data point (image) based on the distance distribution from that point. This results in an adaptive KNN structure for the dataset, which in turn forms the foundation for an adaptive graph construction. We initialized the parameters with $k_{min} = 5$ and $k_{max} = 10$, where $k_{min} = 5$  was chosen as the smallest value that ensured connected graphs across all representations. To assess the robustness of our findings, we further evaluated the results over additional parameter pairs.

\subsection{Ricci Flow Metric Analysis}
\label{ricci-flow-metric}
Ricci flow, originally introduced by Richard S. Hamilton in the context of Riemannian geometry, is a geometric evolution process analogous to heat diffusion. It iteratively updates the metric tensor to smooth out irregularities in the underlying geometry. In this process, regions of positive curvature contract while regions of negative curvature expand. This concept was notably employed by Perelman in his proof of the Poincaré conjecture, where it facilitated the geometric decomposition of 3-manifolds.
The notion of Ricci flow has since been extended to discrete settings, such as graphs. In particular, \cite{ni2019community} proposed a discrete Ricci flow framework in which edge weights evolve based on their corresponding curvature values, specifically, the Ollivier-Ricci curvature. At each iteration, all edge weights are updated simultaneously. Analogous to the continuous case, the discrete Ricci flow shrinks edges with positive curvature and stretches those with negative curvature.
Building on this framework, we apply the discrete Ricci flow process to our graph data, iterating the flow for a fixed number of steps (30 iterations). We initialize all edge weights to one, $w^0_{xy} = w_{xy}$ and $d_{xy}^0 = d_{xy}$ in formula \ref{for:flow}, allowing the final weights to reflect the extent to which each edge must be contracted or expanded based on its curvature. This transformation reveals intrinsic geometric properties of the graph and facilitates geometry-aware comparisons across graphs. 
\subsection{RDMs comparison}
To have a better overview of the comparison between the two viewing conditions, 2D and 3D, across different metrics, we provided the visualisation of the RDM matrices in \figureref{fig:all-RDMs}. In addition, the result of the pair-wise correlation between RDMs, in \figureref{fig:all-RSA}, is also provided.

\begin{figure}[htbp]
\floatconts
  {fig:all-RDMs}
  {\caption{Representational Dissimilarity Matrices (RDMs) of all representations using different metrics for $(k_{\min}, k_{\max}) = (5,10)$ for a) 2D and b) 3D conditions.}}
  {\includegraphics[width=0.85\linewidth]{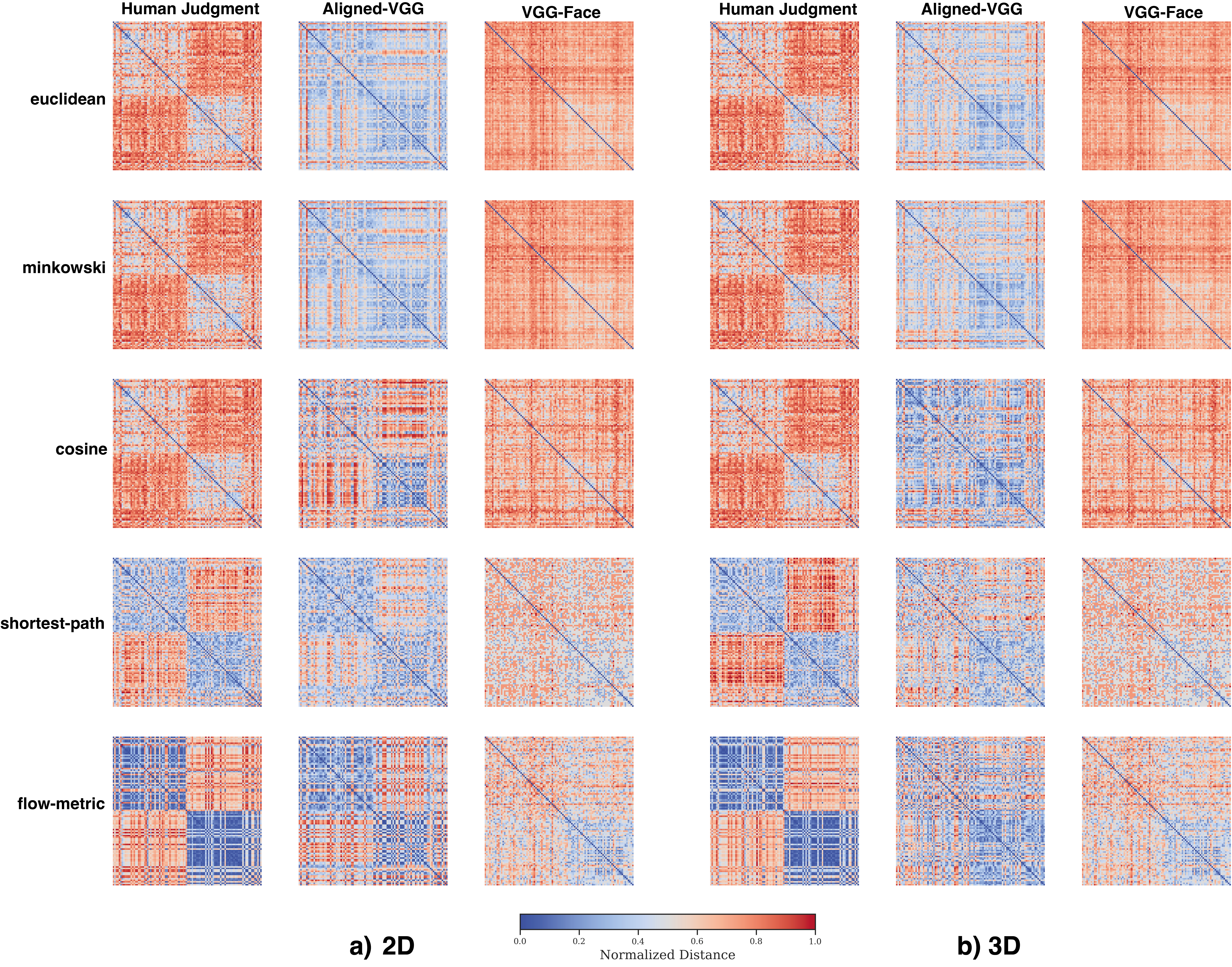}}
\end{figure}

\begin{figure}[htbp]
\floatconts
  {fig:all-RSA}
  {\caption{Correlation of RDMs of all pairs of representations using different metrics with $(k_{\min}, k_{\max}) = (5,10)$ for a) 2D and b) 3D conditions.}}
  {\includegraphics[width=1.0\linewidth]{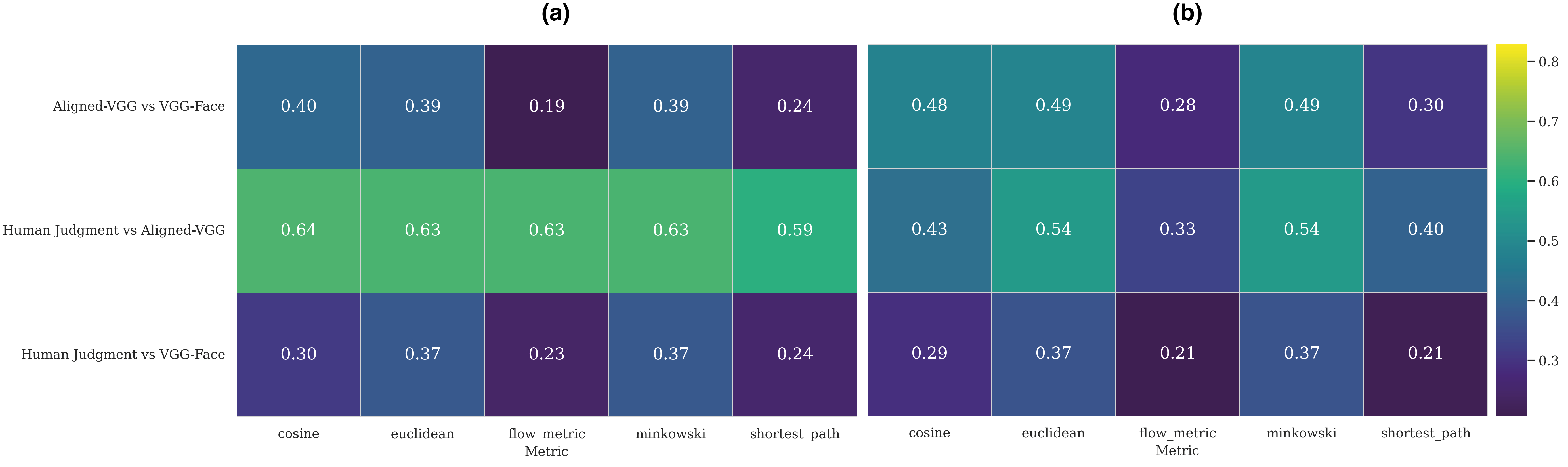}}
\end{figure}
\newpage

\subsection{Synthetic data}
\label{par:torus}
\begin{figure}[htbp]
    \centering
    \includegraphics[width=0.5\linewidth]{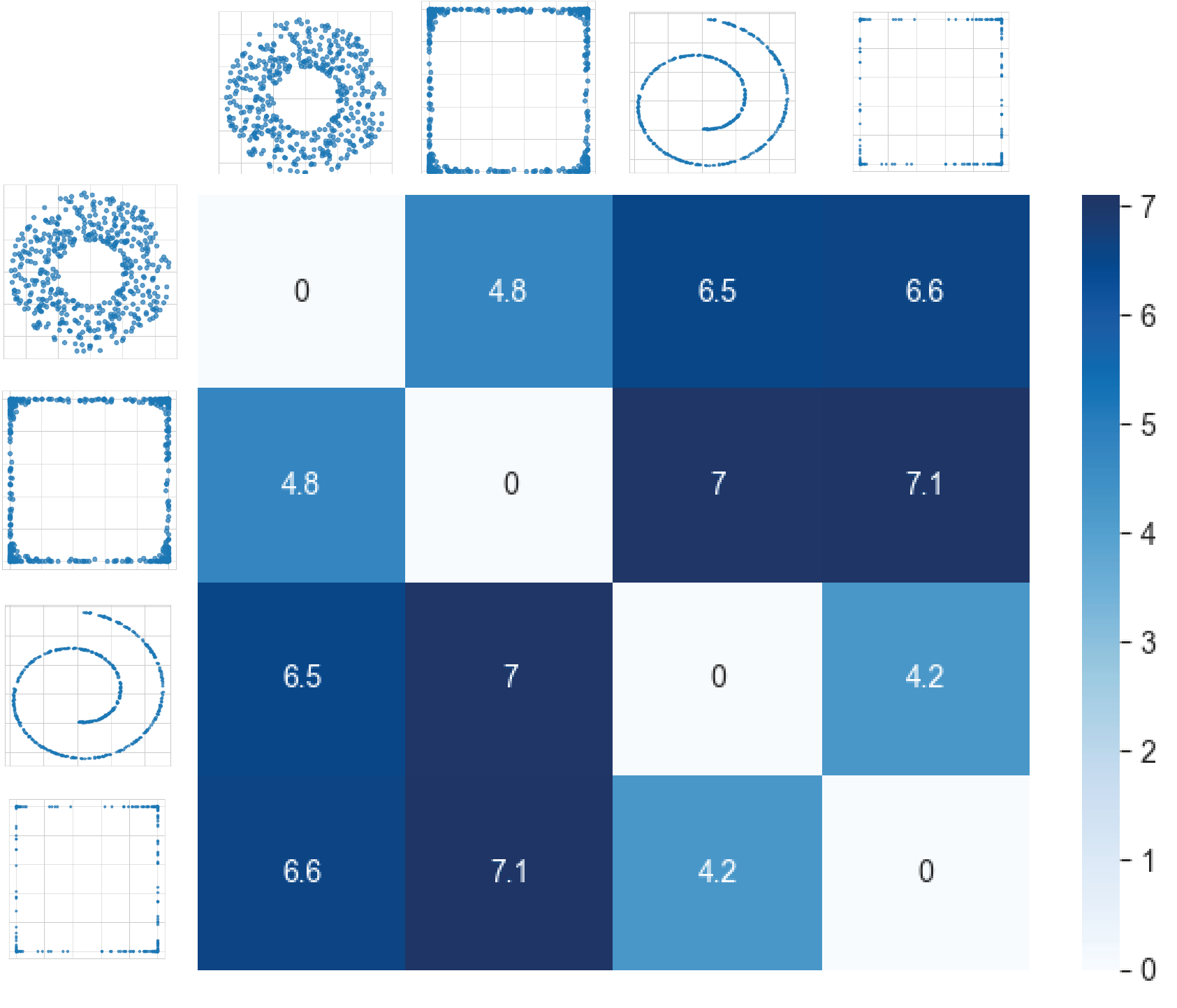}
    \caption{Pairwise heat diffusion distances between the different synthetic datasets. From top to bottom (left to right): 2D torus, transformed 2D torus, 2D swiss roll, transformed 2D swiss roll. Colors indicate heat diffusion distance.}
    \label{fig:syn_data}
\end{figure}

To evaluate whether our representational alignment approach, based on a graph framework, can capture local geometric properties while comparing different representations with varying underlying geometries, we generated several synthetic datasets. Specifically, we created a 2D torus dataset and a dataset resembling a 2D version of a Swiss roll, as illustrated in Figure\ref{fig:syn_data}. To examine the extent to which variation in the underlying geometry can be captured by the heat diffusion distance, we transformed the original data using the sigmoid function $S(x) = \frac{1}{1+e^{-x}}$. The data sets are shown in Figure \ref{fig:syn_data}.
We then constructed a graph representation of each data set as outlined above and computed the pairwise heat diffusion distance. Figure \ref{fig:syn_data} shows the discrepancy captured by the heat diffusion distance, indicating the underlying structural differences between the datasets and their sigmoid compressions. Comparing the distance values within and across data sets shows that the heat diffusion distance could distinguish underlying structures that are more similar within a dataset and its transformation than between datasets or their transforms.

\subsection{Gaussian Mixture Models}
The results of the GMMs for representations by increasing the $k_{\text{min}}$ and $k_{\text{max}}$ for 2D viewing condition are provided in \figureref{fig:Gaussians-image-2D} and for 3D condition in \figureref{fig:Gaussians-image-3D}.

\begin{table}[htbp]
\floatconts
  {tab:gaussian-results}
  {\caption{Results of the GMMs in 2D and 3D conditions.}}
  {\begin{tabular}{lcccc}
  \toprule
  \bfseries \textbf{Model} & \textbf{n-components} & \textbf{BIC} & \textbf{Mean} & \textbf{Covariances}\\
    \midrule
    Human Judgment 2D    & 2  & -277 & $(0.18,-0.08)$ & $(0.01,0.04)$\\
    Human Judgment 3D      & 2 & -309  & $(0.19,-0.10)$ & $(0.01,0.03)$\\
    Aligned-VGG 2D   & 2 & -213 & $(0.21,-0.08)$ & $(0.01,0.04)$\\
    Aligned-VGG 3D   & 1 & -461 & 0.07 & 0.02\\
    VGG-Face        & 1 & -901 & 0.01 & 0.01 \\
  \bottomrule
  \end{tabular}}
\end{table}

\begin{figure}[htbp]
\floatconts
  {fig:Gaussians-image-2D}
  {\caption{GMMs of curvature distributions of all representations in 2D viewing conditions for $(k_{min},k_{max})$= a) (5,15), b) (10,15), and c) (10,20).}}
  {\includegraphics[width=0.7\linewidth]{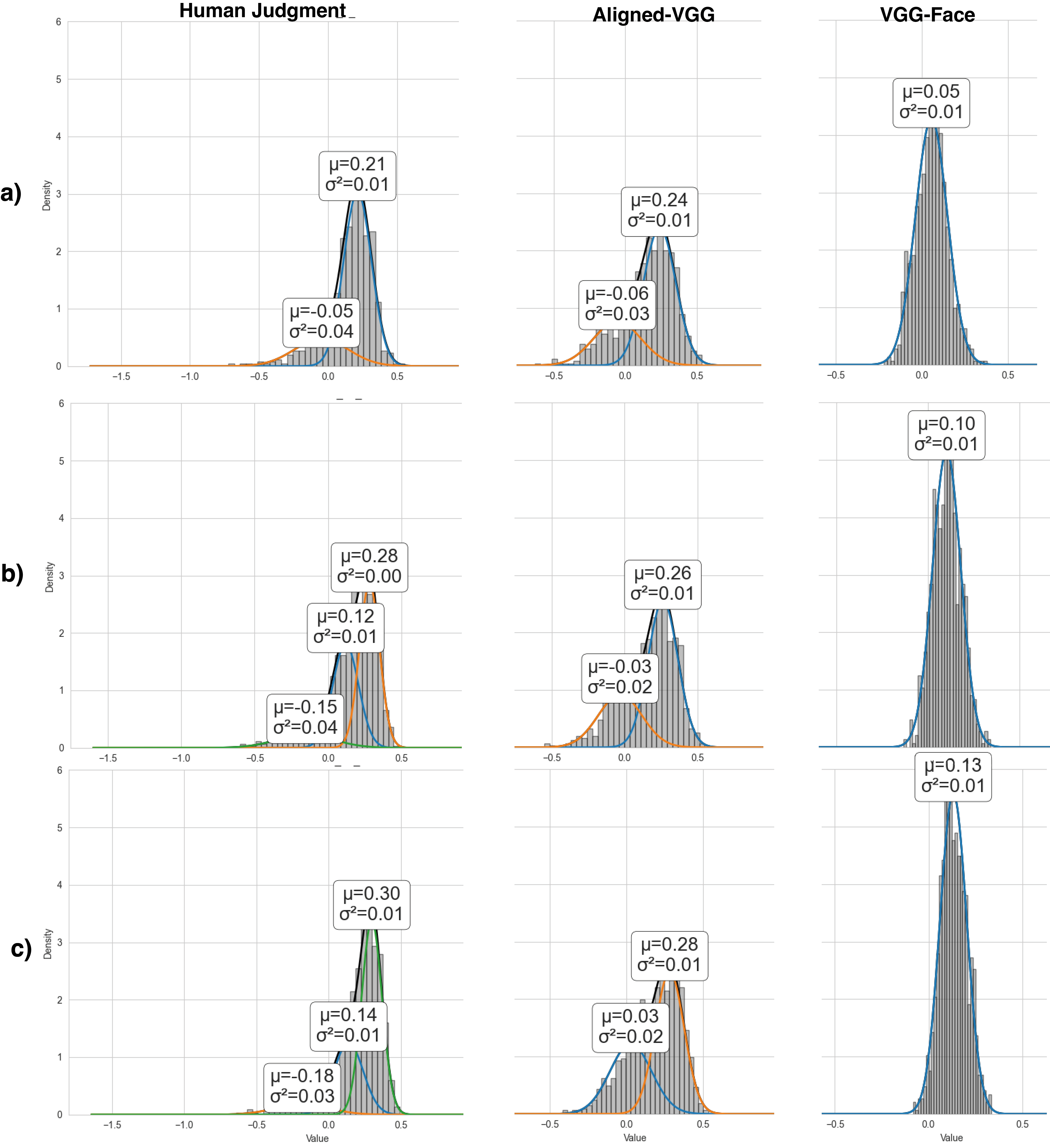}}
\end{figure}

\begin{figure}[htbp]
\floatconts
  {fig:Gaussians-image-3D}
  {\caption{GMMs of curvature distributions of all representations in 3D viewing condition for $(k_{min},k_{max})$= a) (5,15), b) (10,15), and c) (10,20).}}
  {\includegraphics[width=0.7\linewidth]{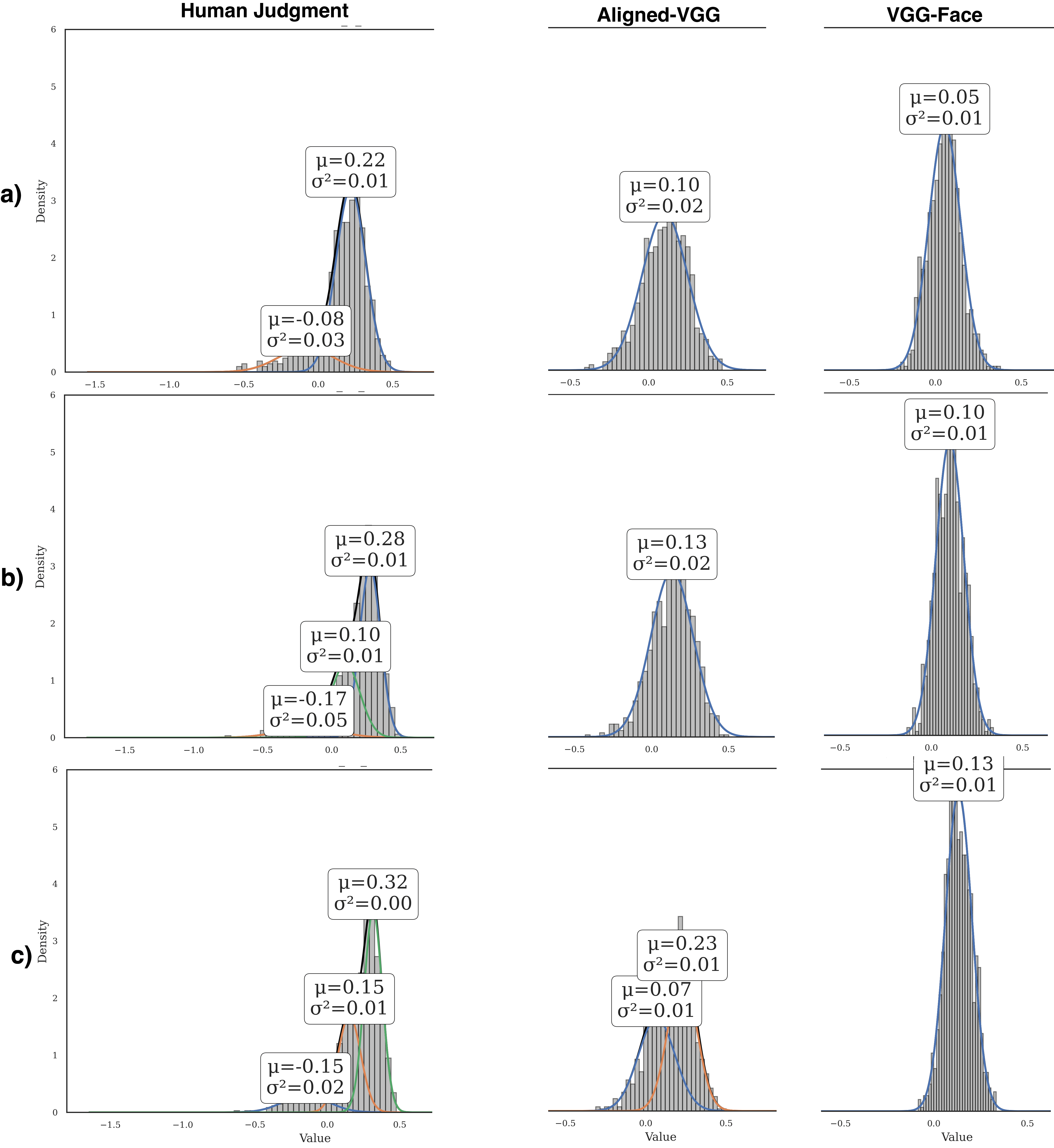}}
\end{figure}
\newpage

\subsection{Heat diffusion distance}
The heatmaps of the Heat diffusion distance for $(k_\text{min},k_{\text{max}}) = (5,15) , (10,20)$ for 2D condition is provided in \figureref{fig:heat-distances-2D} and for 3D is provided in \figureref{fig:heat-distances-3D}.
\begin{figure}[htbp]
\floatconts
  {fig:heat-distances-2D}
  {\caption{Heat diffusion distance results for 2D conditions.}}
  {%
    \subfigure[$k_{\min},k_{\max} = (5,15)$]{\label{fig:image-a}%
      \includegraphics[width=0.4\linewidth]{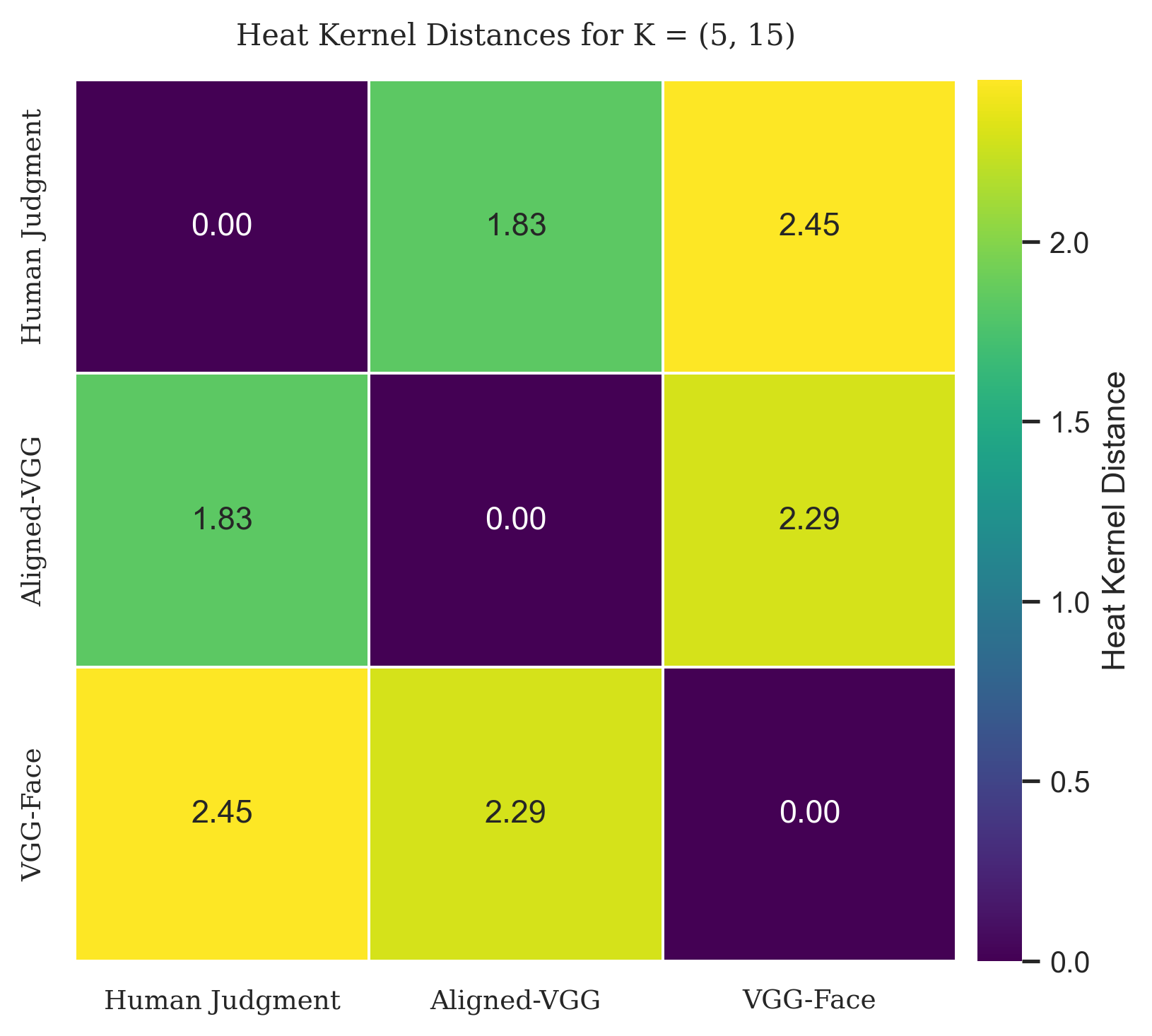}}%
    \qquad
    \subfigure[$k_{\min},k_{\max} = (10,20)$]{\label{fig:image-b}%
      \includegraphics[width=0.4\linewidth]{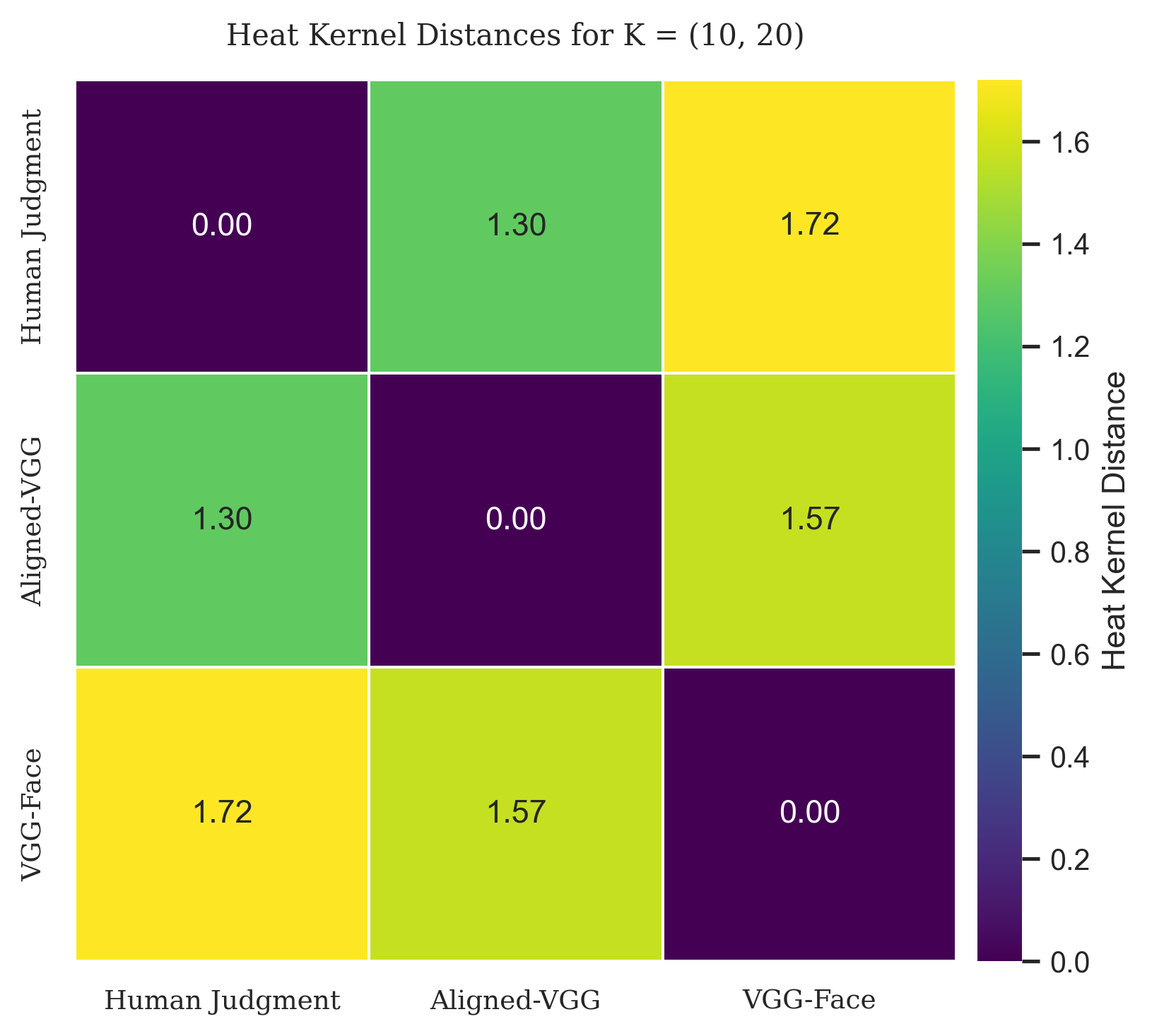}}
  }
\end{figure}

\begin{figure}[htbp]
\floatconts
  {fig:heat-distances-3D}
  {\caption{Heat diffusion distance results for 3D conditions.}}
  {%
    \subfigure[$k_{\min},k_{\max} = (5,15)$]{\label{fig:image-a}%
      \includegraphics[width=0.4\linewidth]{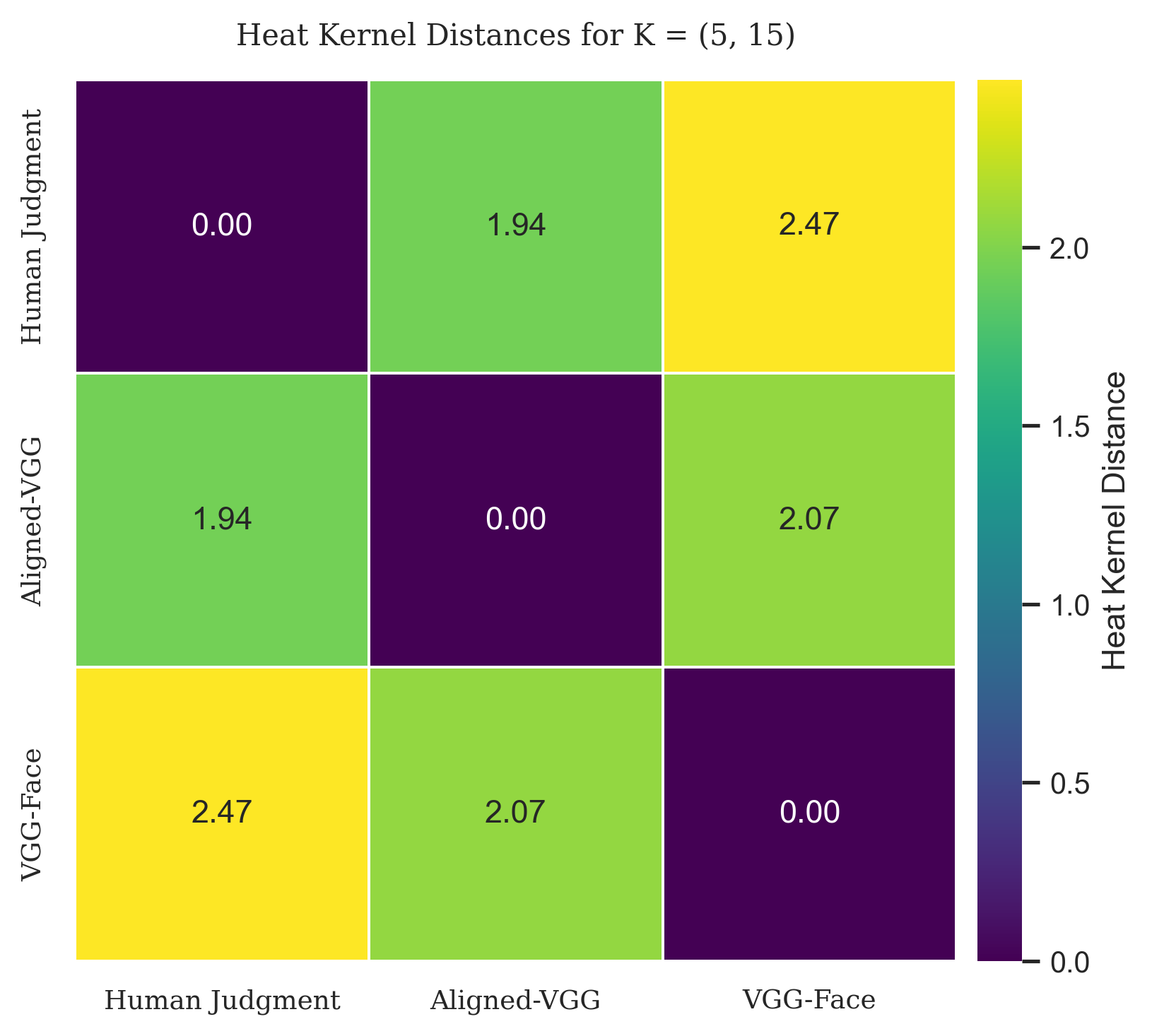}}%
    \qquad
    \subfigure[$k_{\min},k_{\max} = (10,20)$]{\label{fig:image-b}%
      \includegraphics[width=0.4\linewidth]{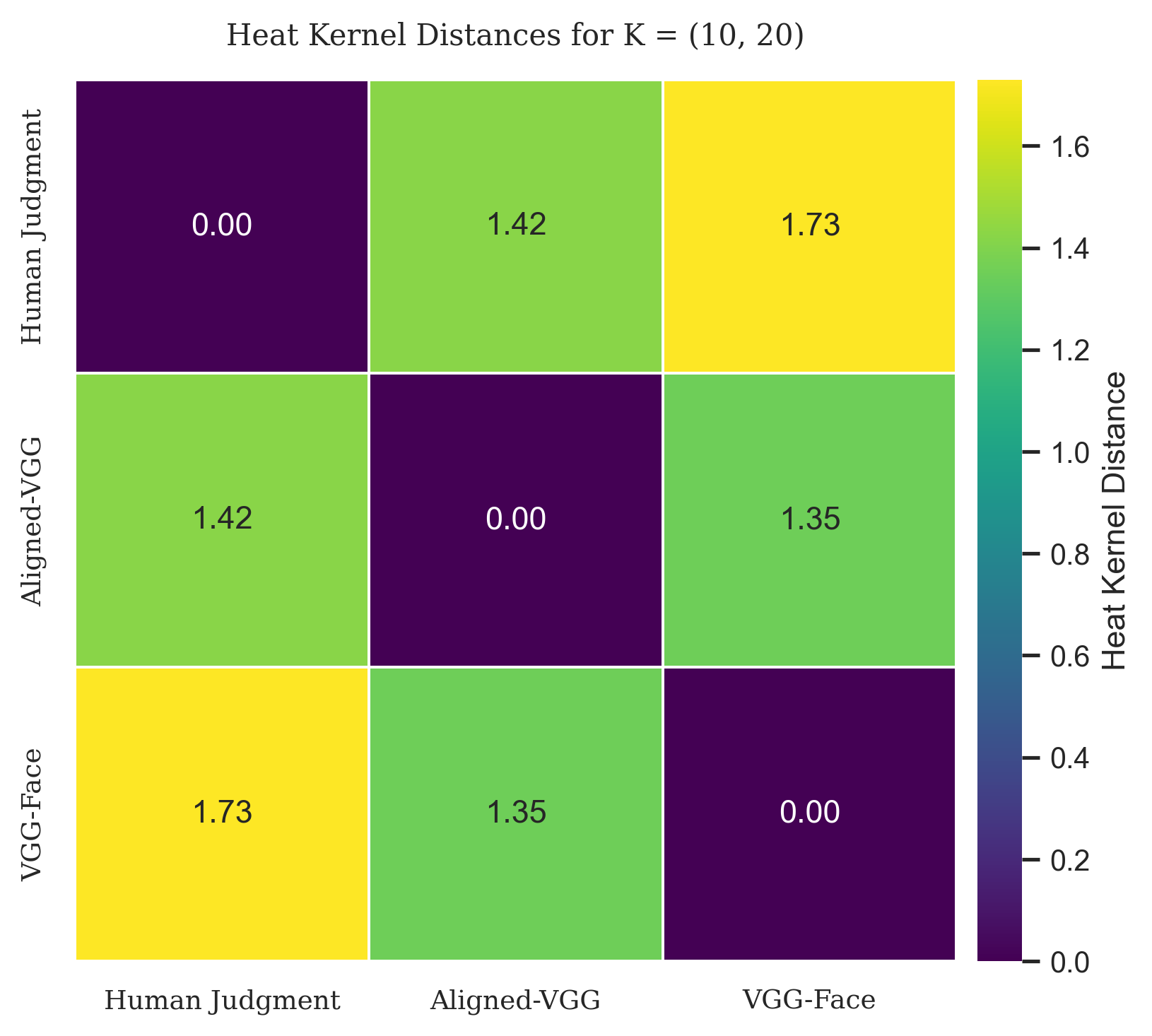}}
  }
\end{figure}
\newpage

\subsection{Wasserstein Distance}

The heatmap results of the Wasserstein distance between the curvature distributions of representations for  $(k_\text{min},k_{\text{max}}) = (5,15), (10,20)$ are provided in 
\figureref{fig:ws-distances-2D} for 2D viewing condition and in \figureref{fig:ws-distances-3D} for 3D viewing condition.

\begin{figure}[htbp]
\floatconts
  {fig:ws-distances-2D}
  {\caption{Wasserstein distance results for 2D condition.}}
  {%
    \subfigure[$k_{\min},k_{\max} = (5,15)$]{\label{fig:i}%
      \includegraphics[width=0.4\linewidth]{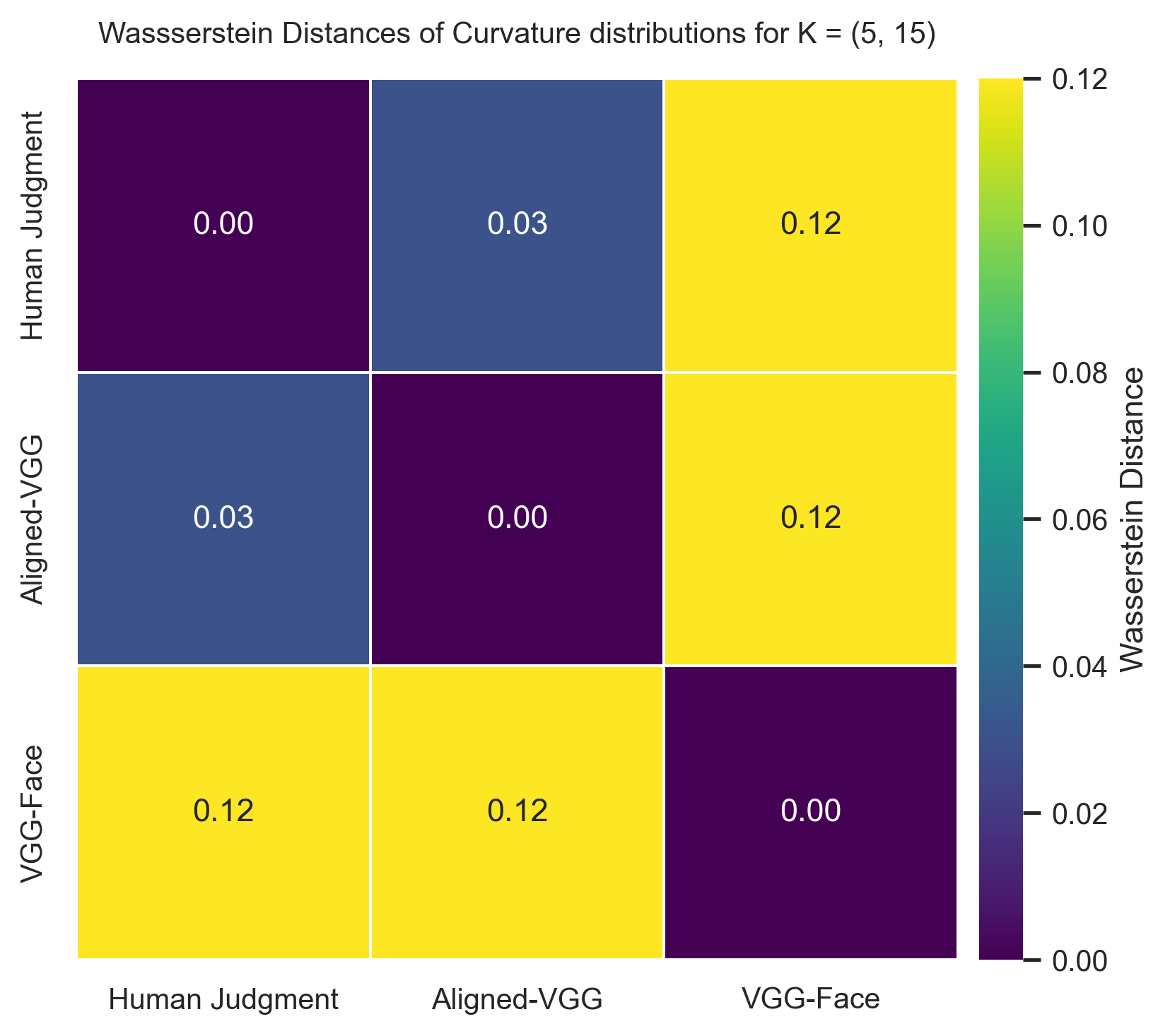}}%
    \qquad
    \subfigure[$k_{\min},k_{\max} = (10,20)$]{\label{fig:image-b}%
      \includegraphics[width=0.4\linewidth]{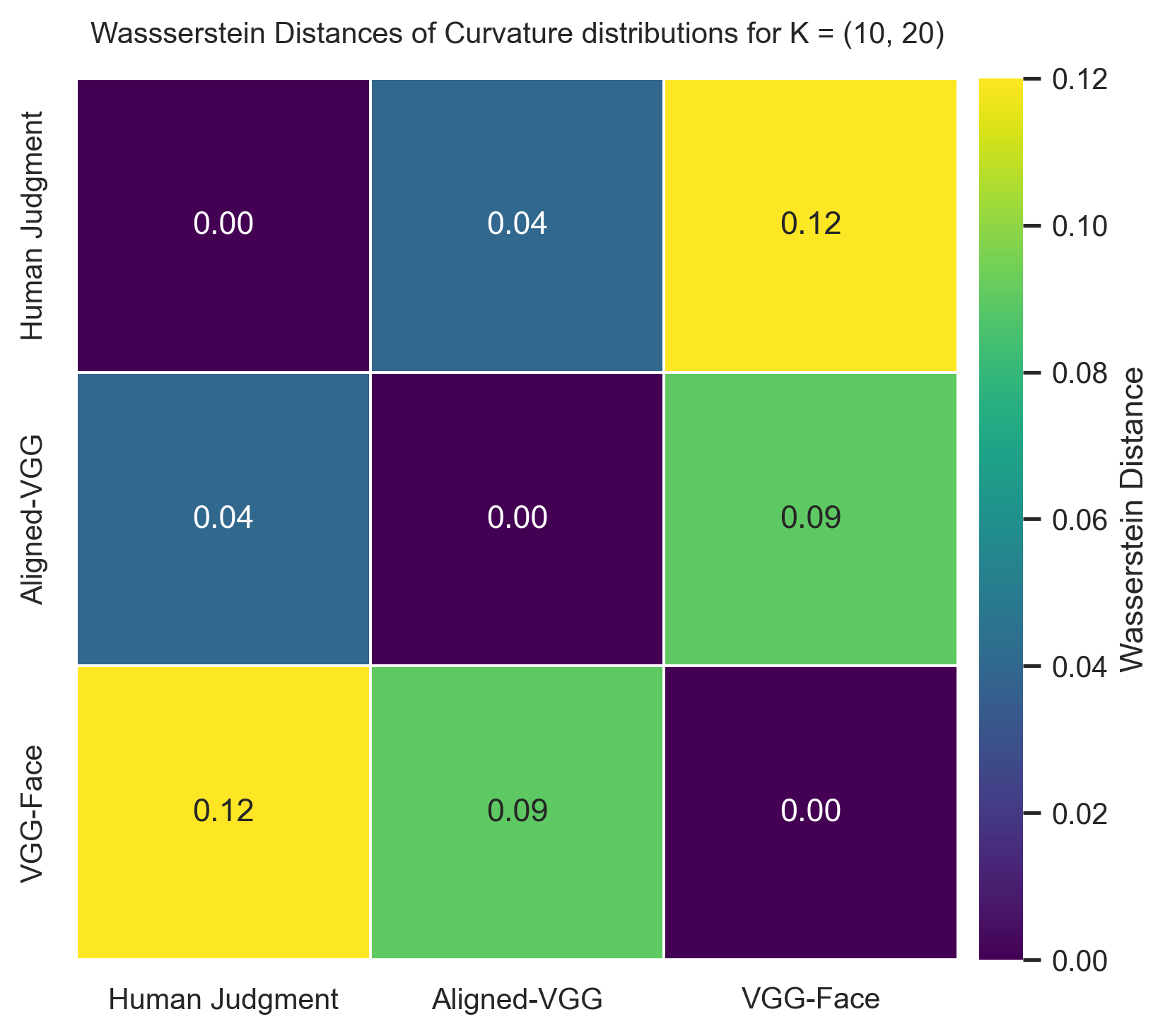}}
  }
\end{figure}

\begin{figure}[htbp]
\floatconts
  {fig:ws-distances-3D}
  {\caption{Wasserstein distance results for 3D condition.}}
  {%
    \subfigure[$k_{\min},k_{\max} = (5,15)$]{\label{fig:i}%
      \includegraphics[width=0.4\linewidth]{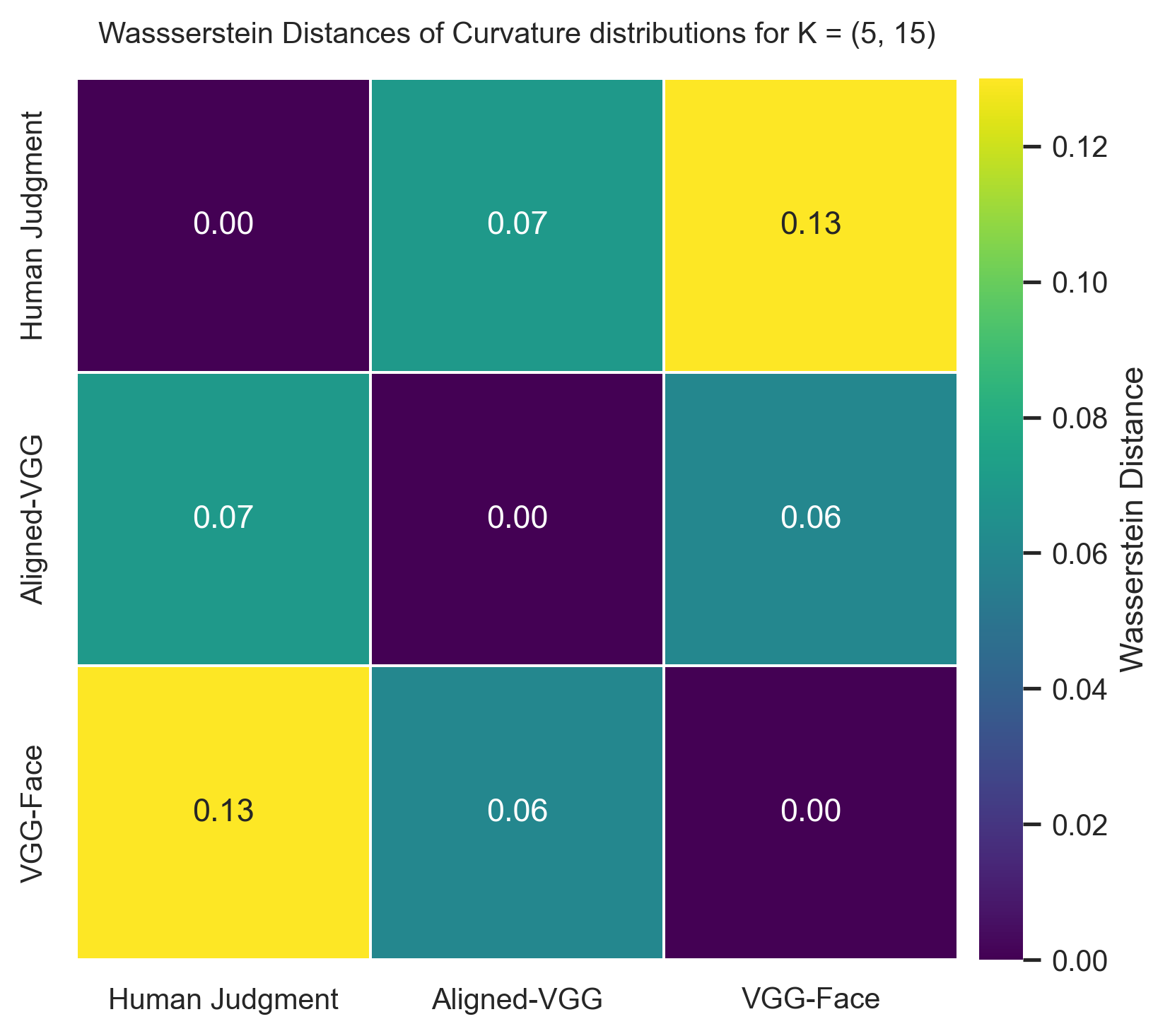}}%
    \qquad
    \subfigure[$k_{\min},k_{\max} = (10,20)$]{\label{fig:image-b}%
      \includegraphics[width=0.4\linewidth]{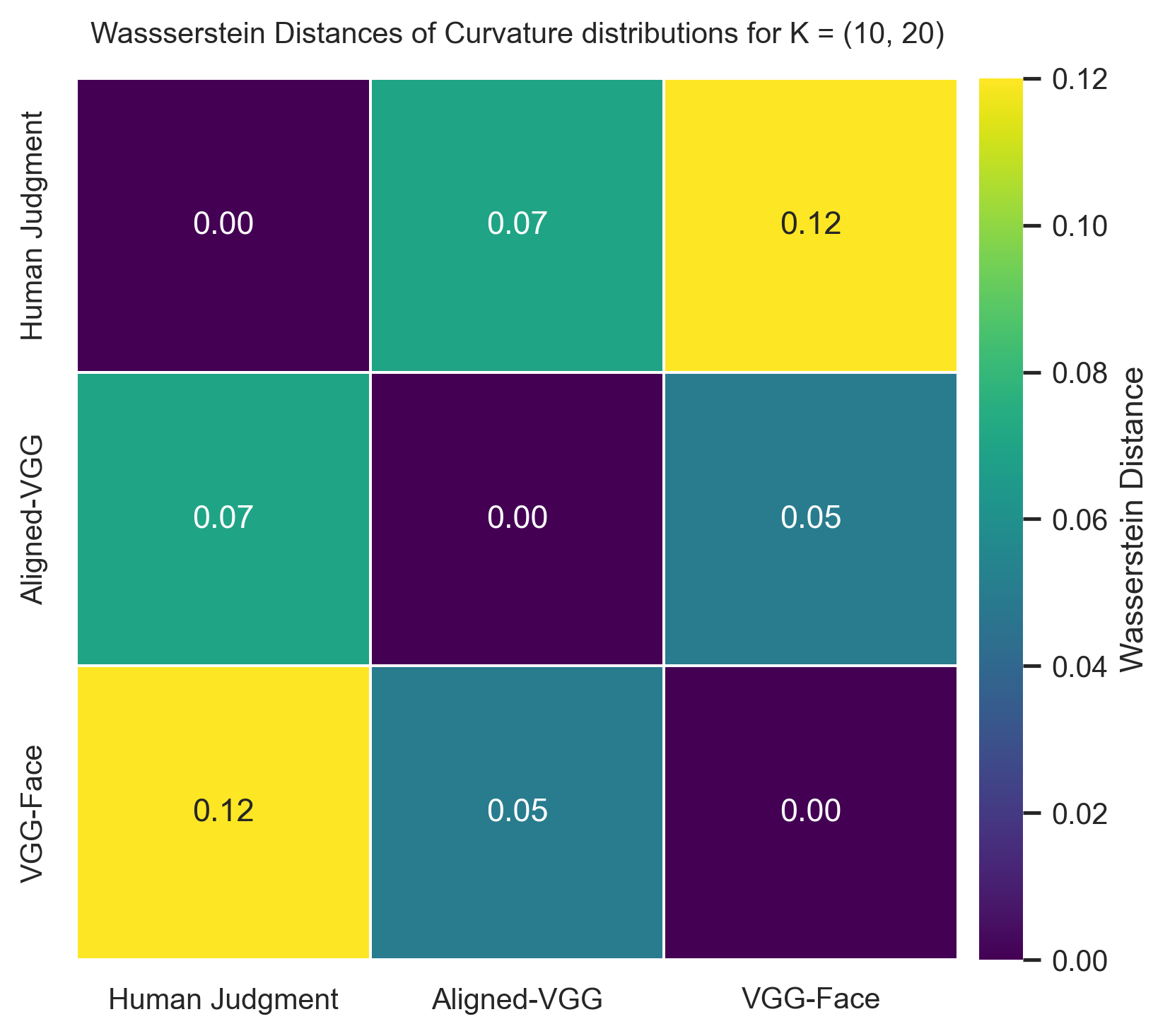}}
  }
\end{figure}

\subsection{Representational Profile Analysis}
\begin{figure}[htbp]
    \centering
    \includegraphics[width=1.0
    \linewidth]{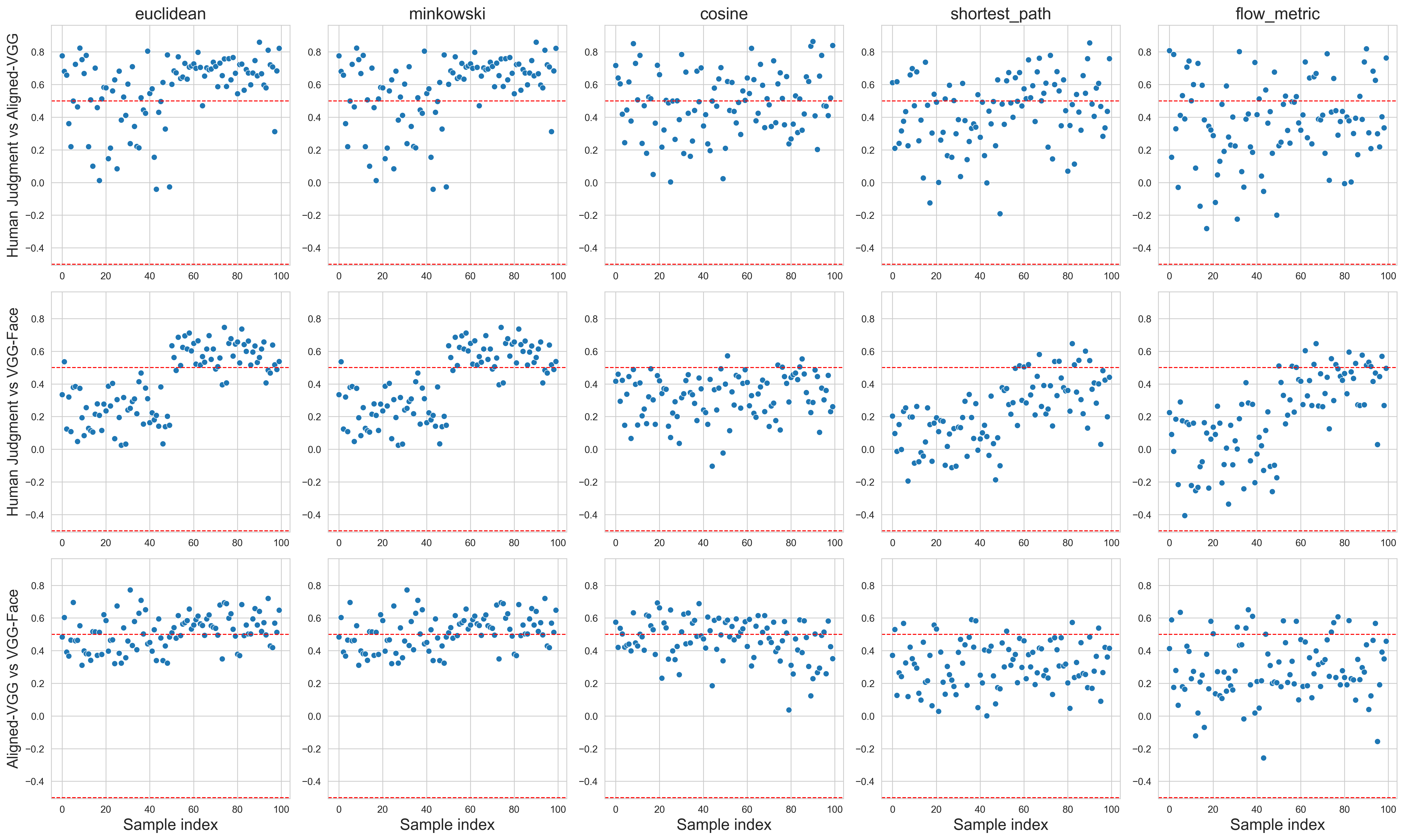}
    \caption{Point-level correlation of the data points between pairs of RDM matrices computed by different metrics (euclidean, shortest-path, flow-metric,minkowski, and cosine from the left, respectively) for $(k_{min}, k_{max}) = (5,10)$ for \textbf{3D} condition. Red dashed lines show r=0.5 (up), and r=-0.5 (down).}
    \label{fig:row-wise-3D}
\end{figure}

\begin{figure}[htbp]
    \centering
    \includegraphics[width=1.0\linewidth]{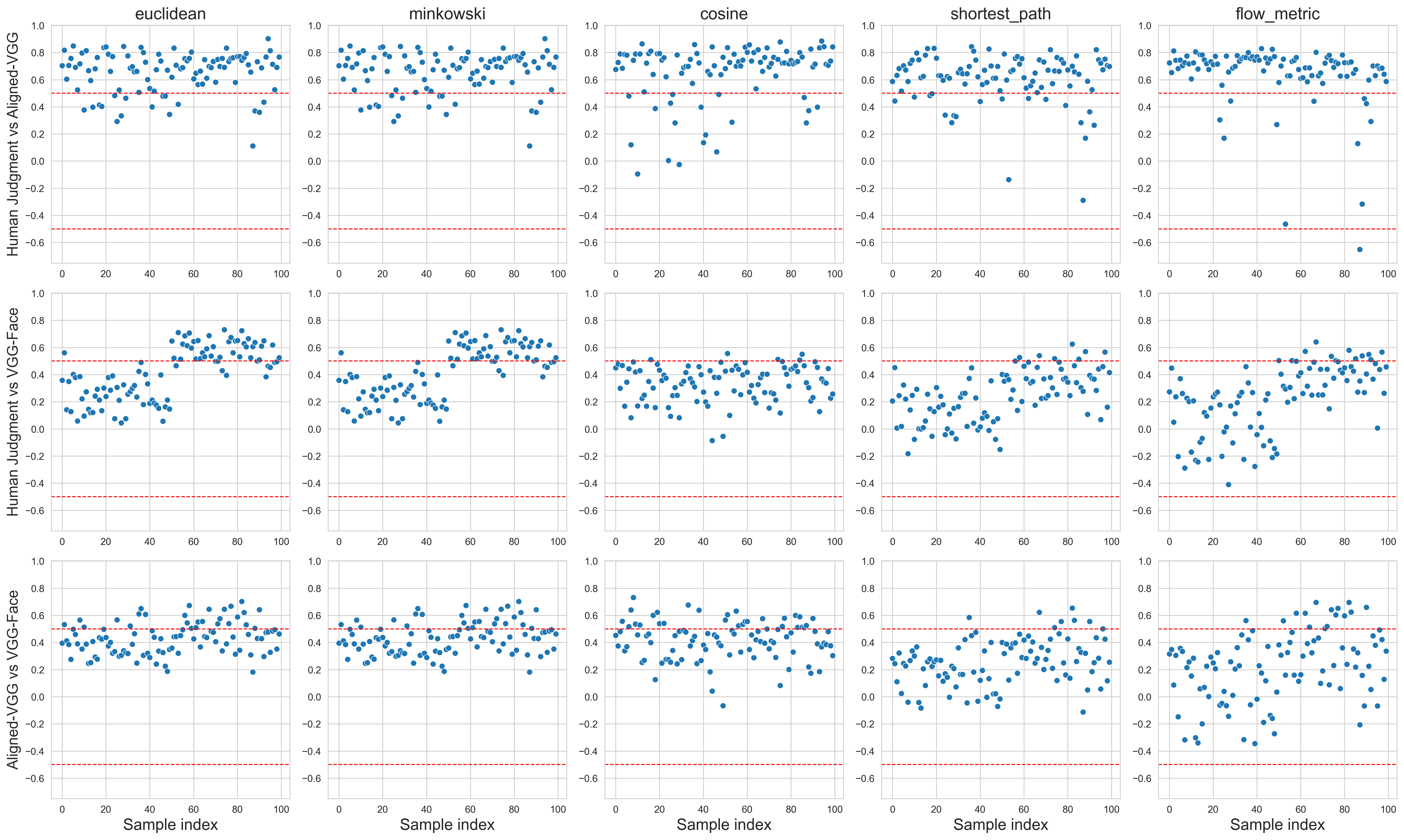}
    \caption{Point-level correlation of the data points between pairs of RDM matrices computed by different distance metrics (euclidean, shortest-path, flow-metric,minkowski, and cosine from the left, respectively) for $(k_{min}, k_{max}) = (5,10)$ for \textbf{2D condition}. Red dashed lines show r=0.5 (up), and r=-0.5 (down).}
    \label{fig:row-wise-2D}
\end{figure}

To gain a more fine-grained understanding of how image representations vary and relate across representational spaces, we conducted an additional analysis. For each image, in a given representational space, we extracted its distance profile,  any point in the figures, corresponding to the relevant row of its RDM, and correlated it with the same row from the RDMs of other representations, using multiple distance metrics. 
In \textbf{3D condition} \figureref{fig:row-wise-3D}, Alignment becomes heterogeneous, splitting into two clusters by gender: male images maintain higher correlations, while female images show weaker alignment, especially between Human Judgment and VGG-based representations under the Euclidean metric. Geometry-aware metrics (shortest path, flow-metric) further reduce and scatter correlations, indicating that increased spatial complexity disrupts geometric preservation. In \textbf{2D condition}, \figureref{fig:row-wise-2D}, Human Judgment and Aligned-VGG representations show consistently high correlations across metrics.\\
This pattern reinforces our earlier findings: in \textbf{2D}, alignment faithfully projects human perceptual geometry into representational space, yielding an arrangement of data points that closely mirrors human face‑perception averages. VGG-Face, by comparison, adopts a distinct geometric layout. In \textbf{3D}, however, human judgment cannot be mapped onto the underlying geometry of either aligned or unaligned network representations. These results suggest that alignment effectively preserves task‑relevant geometric structure only in two dimensions.

\subsection{Community metrics results}
We analyzed the structural properties of the processed graphs, as summarized in Table \ref{tab:results}. In particular, we calculated \textit{conductance} (community separation by the between-and-within edge ratio), \textit{internal edge density} (edges within the community vs. maximum possible), \textit{modularity} (community strength by comparing edge densities with a random graph), and \textit{average embeddedness} (number of shared neighbors for pairs of nodes within a community). These results clearly show that the Human Judgment graph and the graph representation of the Aligned-VGG have very similar properties in terms of community structure. In contrast, the Original VGG-Face, despite being trained on face images, does not show a similarly high degree of structure. Going from a 2D viewing condition to 3D, congruence to Fig \ref{fig:ricci-flow}, it is clear that  Aligned-VGG has properties more similar to VGG-Face rather than Human Judgment, in contrast to the 2D condition.

\begin{table*}[htbp]
\caption{Comparison of graph structure based on KL-divergence (KLD) between edge curvature distributions and community metrics derived from Ricc flow: \textit{Conductance}, \textit{Internal edge density} (IED), \textit{Modularity}, and \textit{Average embeddedness} (AE).}
\label{tab:results}
\vskip 0.15in
\begin{center}
\begin{small}
\begin{sc}
\begin{tabular}{lccccr}
\toprule
Graph  & Communities & Conductance & IED & Modularity & AE \\
\midrule
Human Judgment 2D & 7 & 0.23 & 0.60 & 34.73 & 0.78 \\
Human Judgment 3D & 8 & 0.31 & 0.68 & 31.54 & 0.69 \\
Aligned-VGG 2D & 6 & 0.20 & 0.52 & 35.98 & 0.80 \\
Aligned-VGG 3D & 6 & 0.34 & 0.57 & 21.57 & 0.68 \\
VGG-Face & 3 & 0.33 & 0.32 & 11.75 & 0.69 \\
\bottomrule
\end{tabular}
\end{sc}
\end{small}
\end{center}
\vskip -0.1in
\end{table*}
\newpage

\subsection{Dimensionality Reduction Visualisation of Embeddings}
To visualize the embedding spaces of both native and behaviorally aligned networks under 2D and 3D viewing conditions, we applied three dimensionality reduction techniques: PCA, t-SNE, and UMAP. \figureref{fig:hum-aligned-2d} presents the Human-Aligned network in the 2D condition, \figureref{fig:hum-aligned-3d} shows the same network under the 3D condition, and \figureref{fig:orig-vgg} depicts the baseline VGG-Face model. In each visualization, male and female image embeddings are distinguished by color.

\begin{figure}[htbp]
\floatconts
  {fig:hum-aligned-2d}
  {\caption{2D Visualisation of the embedding vectors for Human-Aligned network in 2D condition. Colors show female and male images.}}
  {%
    \subfigure[PCA]{\label{Neu}%
      \includegraphics[width=0.25\linewidth]{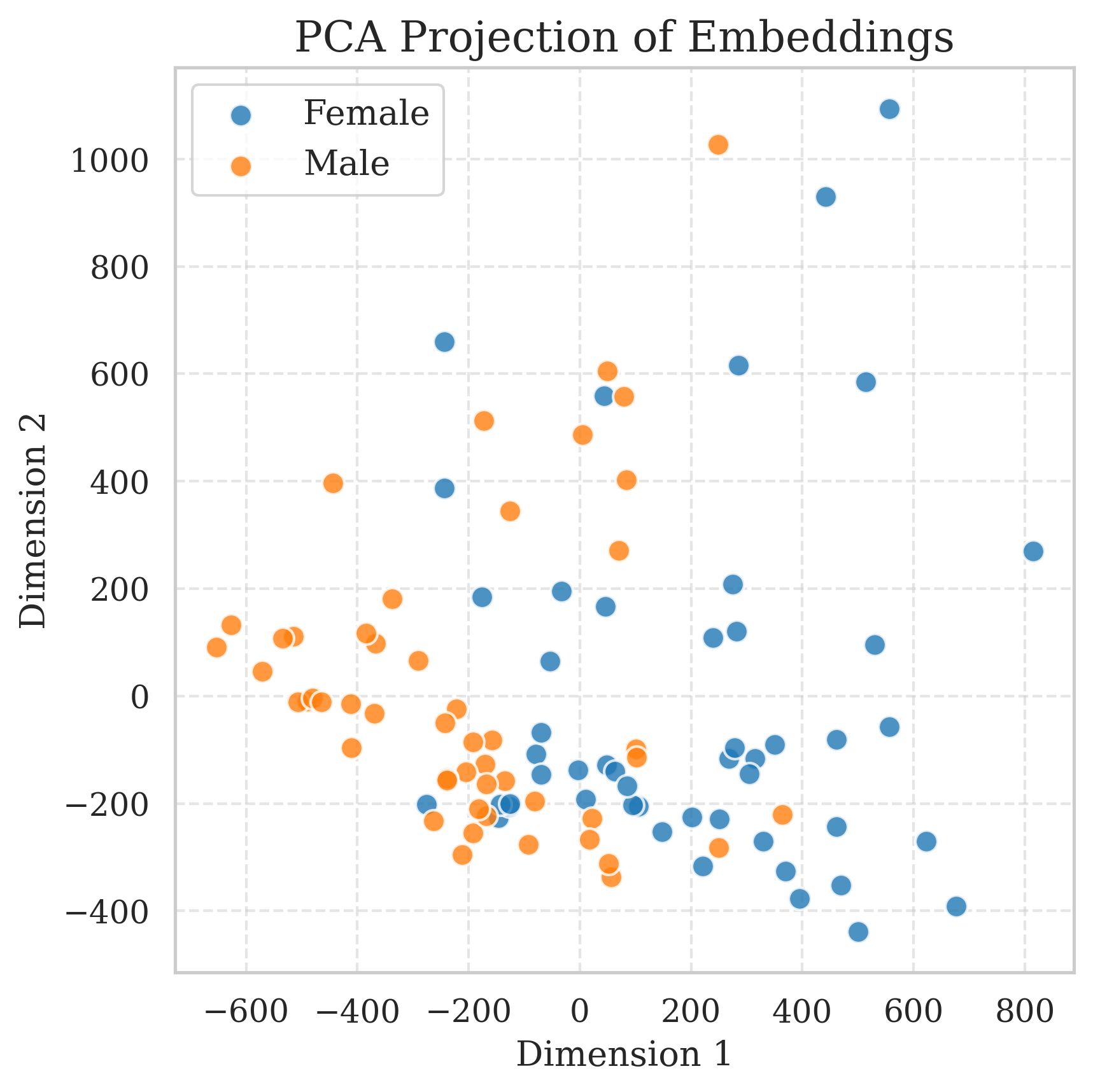}}%
    \qquad
    \subfigure[t-SNE]{\label{fig:image-b}%
      \includegraphics[width=0.25\linewidth]{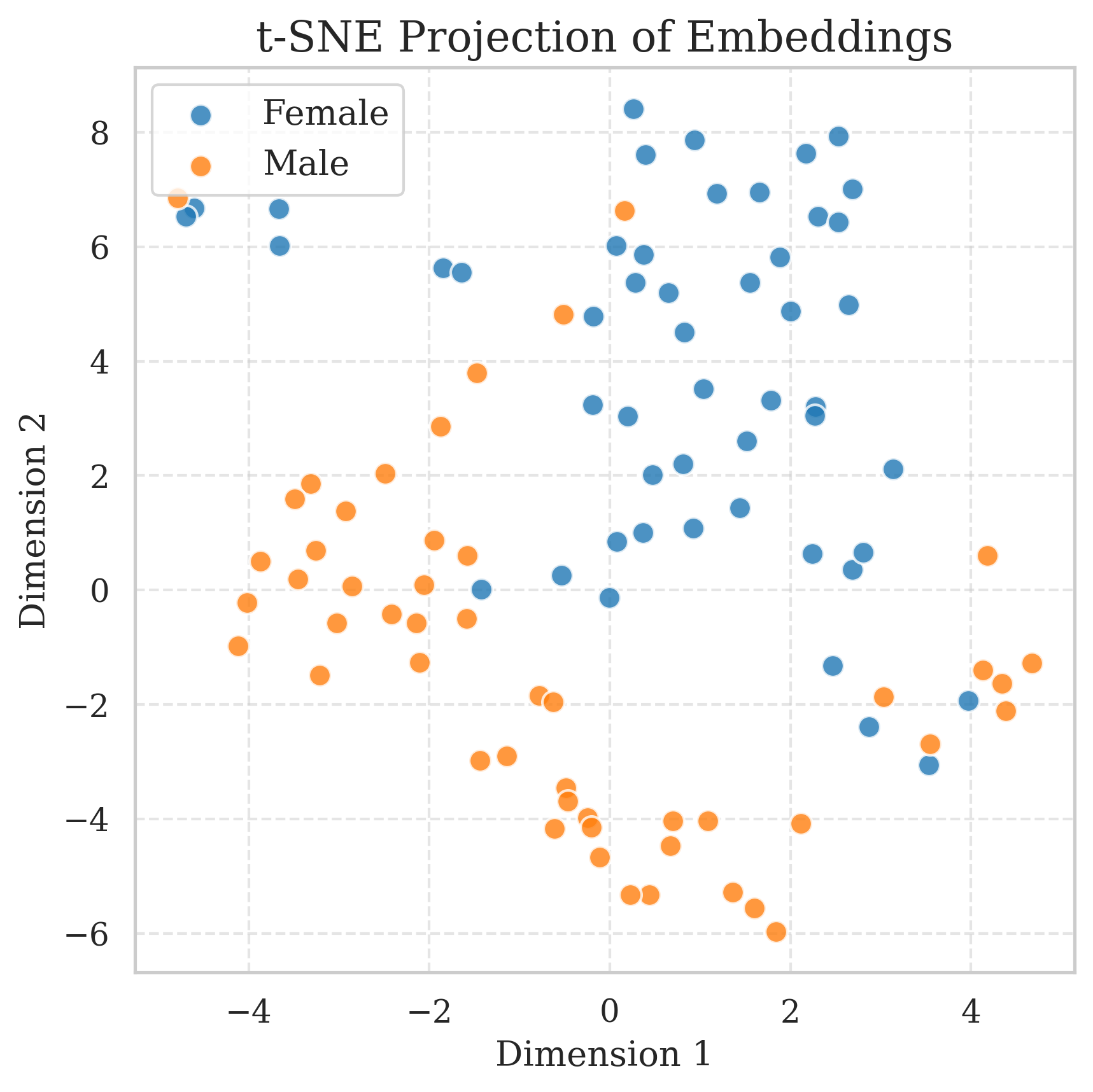}}
      \qquad
    \subfigure[UMAP]{\label{fig:image-b}%
      \includegraphics[width=0.25\linewidth]{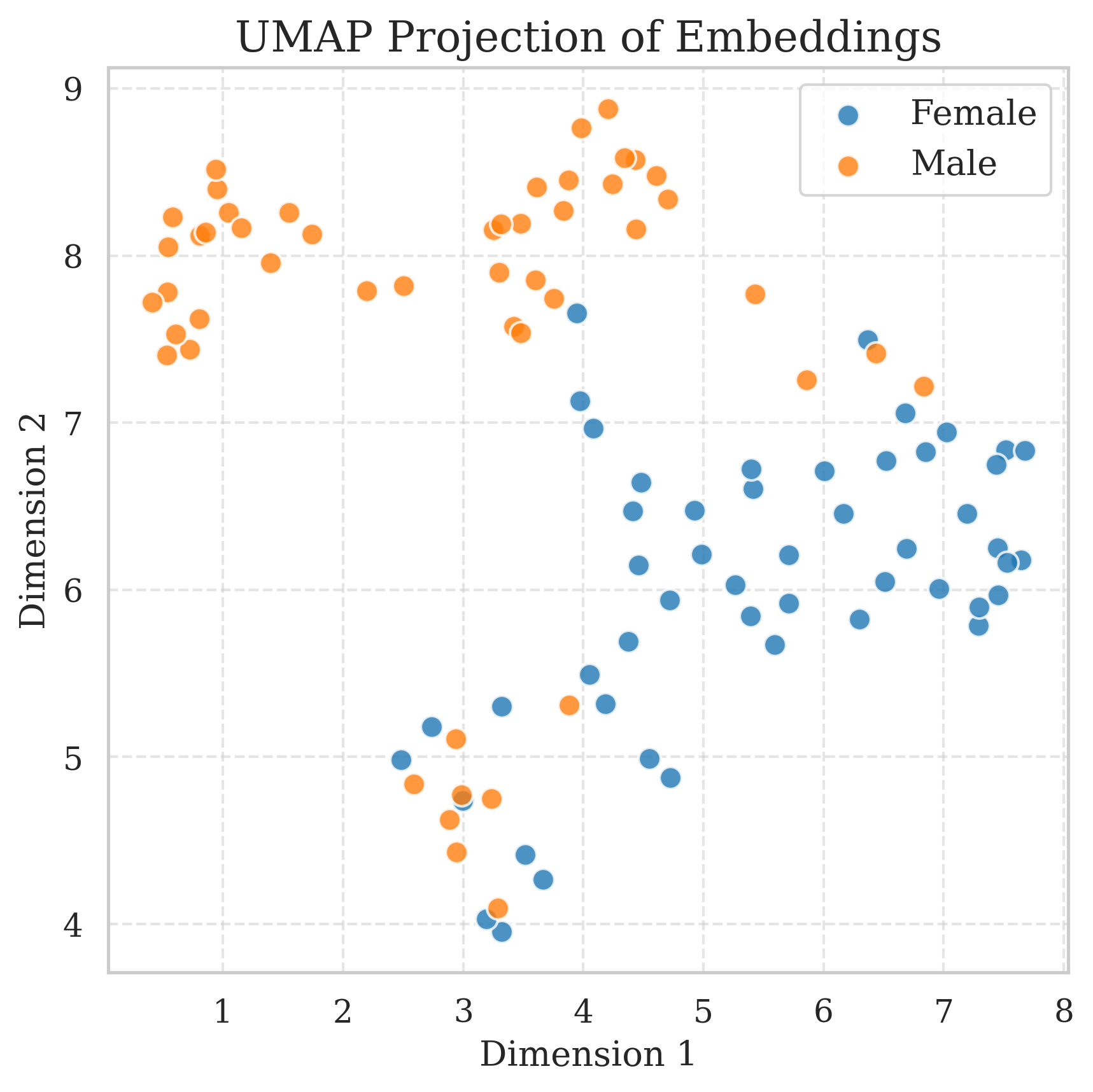}}
  }
\end{figure}

\begin{figure}[htbp]
\floatconts
  {fig:hum-aligned-3d}
  {\caption{2D Visualisation of the embedding vectors for Human-Aligned network in 3D condition. Colors show female and male images.}}
  {%
    \subfigure[PCA]{\label{Neu}%
      \includegraphics[width=0.25\linewidth]{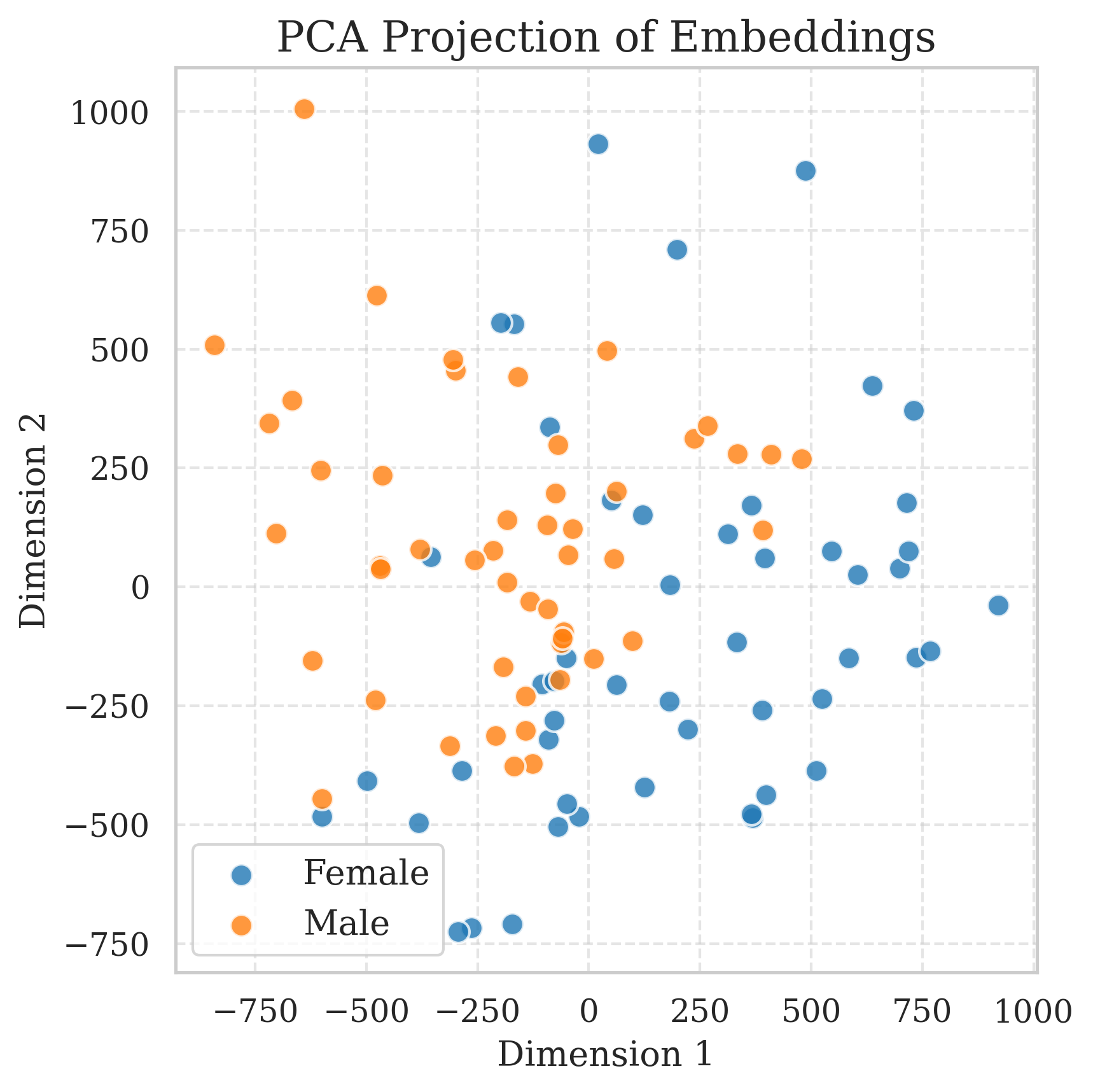}}%
    \qquad
    \subfigure[t-SNE]{\label{fig:image-b}%
      \includegraphics[width=0.25\linewidth]{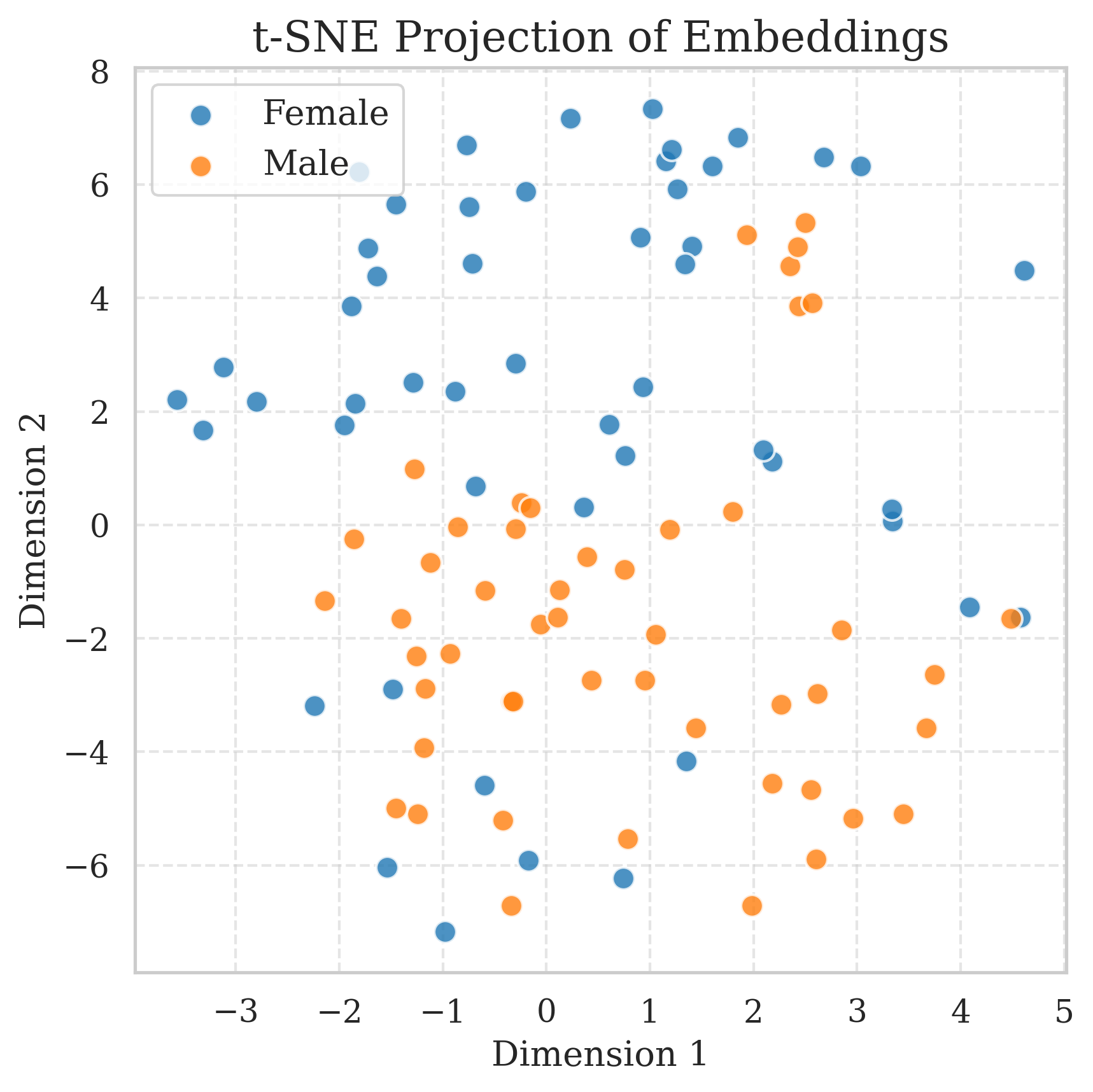}}
      \qquad
    \subfigure[UMAP]{\label{fig:image-b}%
      \includegraphics[width=0.25\linewidth]{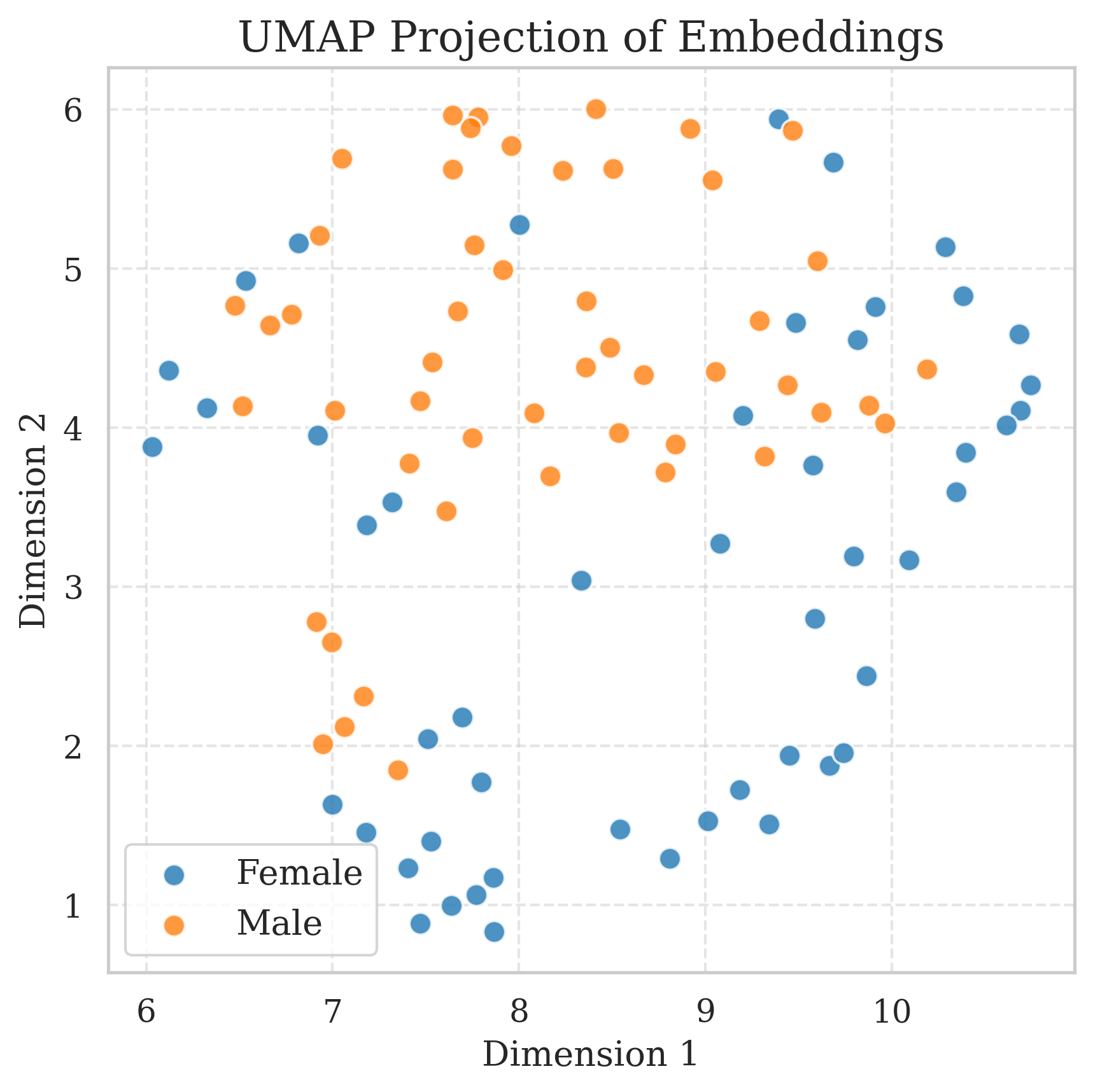}}
  }
\end{figure}

\begin{figure}[htbp]
\floatconts
  {fig:orig-vgg}
  {\caption{2D Visualisation of the embedding vectors for the VGG-Face. Colors show female and male images.}}
  {%
    \subfigure[PCA]{\label{Neu}%
      \includegraphics[width=0.25\linewidth]{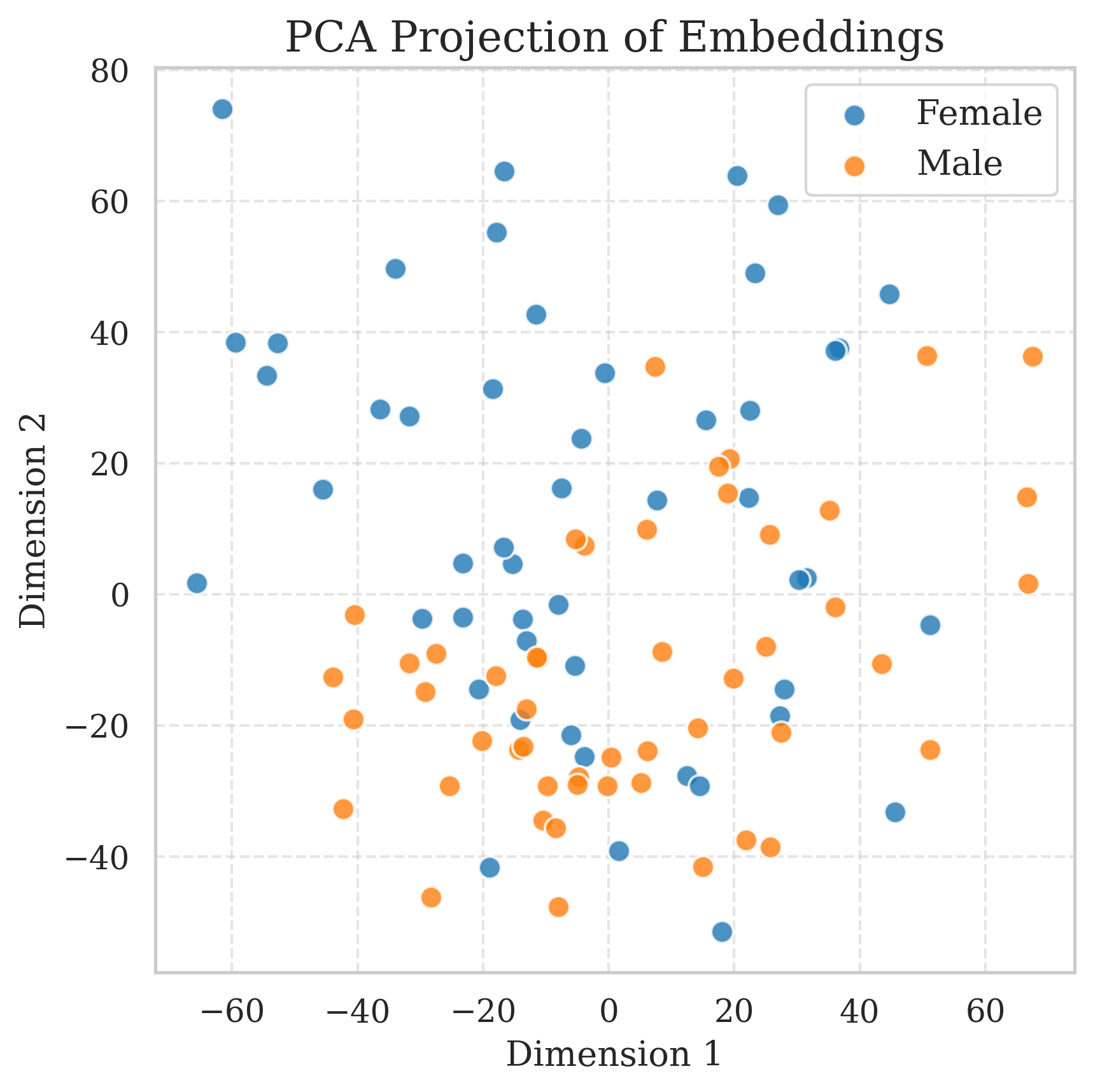}}%
    \qquad
    \subfigure[t-SNE]{\label{fig:image-b}%
      \includegraphics[width=0.25\linewidth]{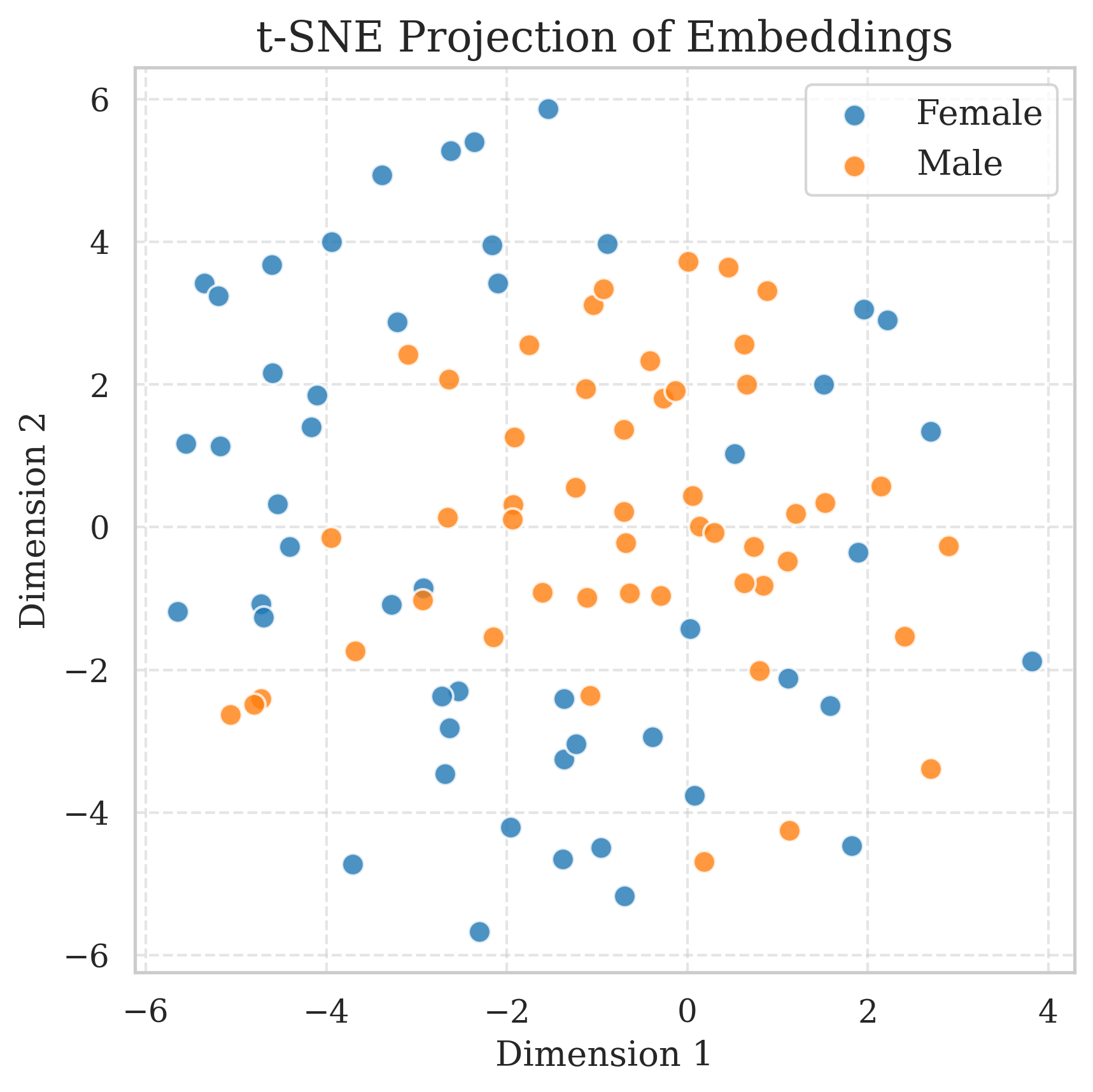}}
      \qquad
    \subfigure[UMAP]{\label{fig:image-b}%
      \includegraphics[width=0.25\linewidth]{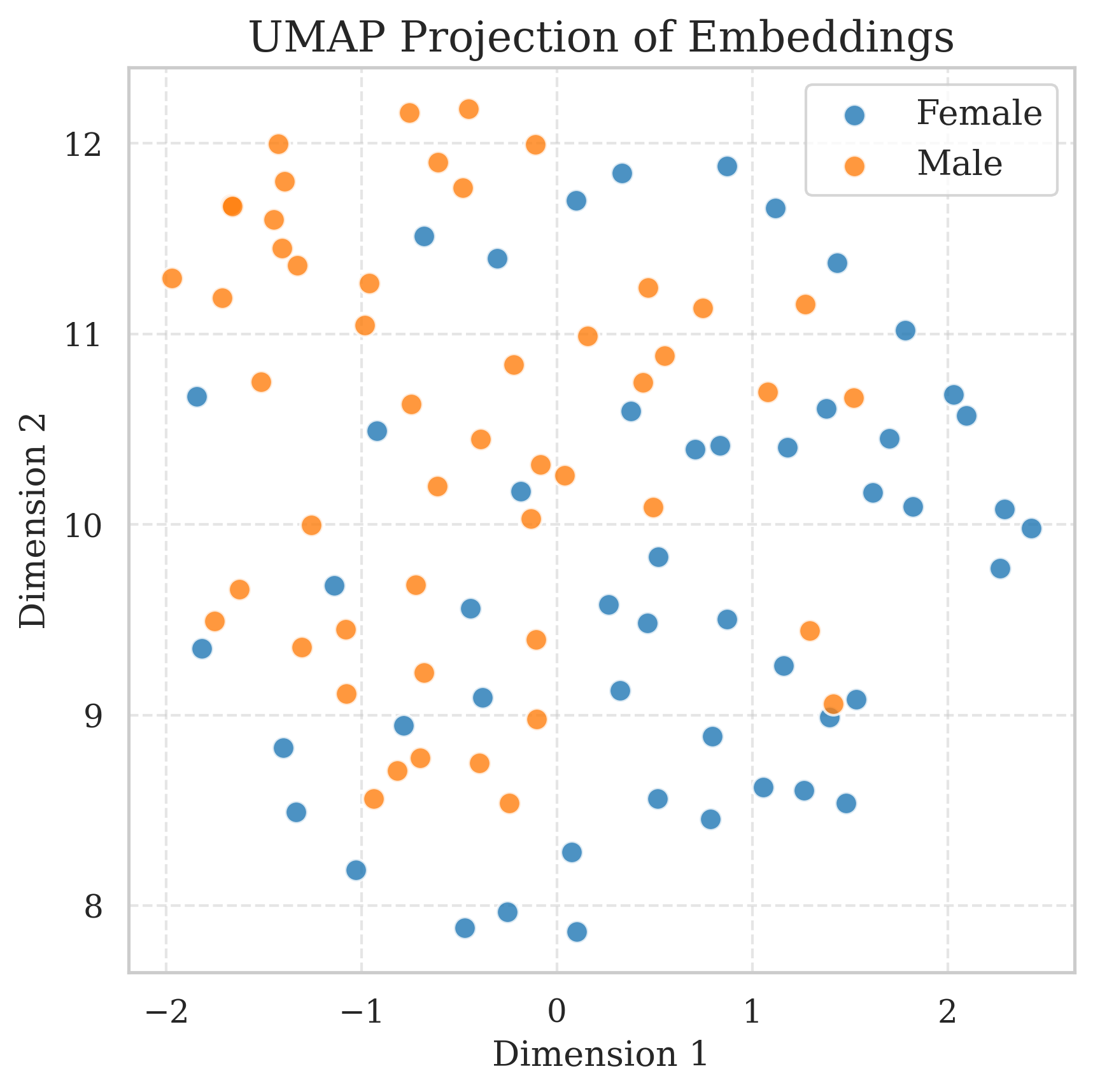}}
  }
\end{figure}

\end{document}